%% file: 0_main.tex
\documentclass[lettersize,journal]{IEEEtran}
\usepackage{amsmath}
\usepackage{algorithmic}
\usepackage{subcaption}
\usepackage{algorithm}
\usepackage{array}
\usepackage{textcomp}
\usepackage{siunitx}
\usepackage{xcolor}
\usepackage{amsfonts} 
\usepackage{multicol}
\usepackage{booktabs}
\usepackage{tabularx}
\usepackage{multirow}
\usepackage{hyperref}
\usepackage{xspace}
 \usepackage{url}
\usepackage{stfloats}
\usepackage{url}
\usepackage{verbatim}
\usepackage{enumerate}
\usepackage{graphicx}
\usepackage{cite}
\usepackage{verbatim}
\usepackage{caption}
\usepackage{ragged2e}
\newcommand{\alberto}[1]{\textcolor{blue}{#1}}

\newcommand{\stanh}{\emph{STanH\,}}

\newcolumntype{P}[1]{>{\centering\arraybackslash}p{#1}}
\newcolumntype{M}[1]{>{\centering\arraybackslash}m{#1}}
\newcolumntype{C}[1]{>{\centering\arraybackslash}m{#1}}

\DeclareMathOperator{\Stanh}{STanH}
\begin{document}

\title{\stanh: Parametric Quantization for \\ Variable Rate Learned Image Compression}

\author{Alberto Presta~\IEEEmembership{Student Member, IEEE}, \and
        Enzo Tartaglione~\IEEEmembership{Member, IEEE},\\ \and Attilio Fiandrotti~\IEEEmembership{Senior Member, IEEE}, \and Marco Grangetto~\IEEEmembership{Senior Member, IEEE} 
}

\markboth{IEEE Transactions on Image Processing, submitted March~2024}%
{Shell \MakeLowercase{\textit{et al.}}: A Sample Article Using IEEEtran.cls for IEEE Journals  }



\maketitle

\begin{abstract}
In end-to-end learned image compression, encoder and decoder are jointly trained to minimize a $\boldsymbol{R} \boldsymbol{+} \boldsymbol{\lambda} \boldsymbol{D}$ cost function, where $\boldsymbol{\lambda}$ controls the trade-off between rate of the quantized latent representation and image quality.
Unfortunately, a distinct encoder-decoder pair with millions of parameters must be trained for each $\boldsymbol{\lambda}$, hence the need to switch encoders and to store multiple encoders and decoders on the user device for every target rate.
This paper proposes to exploit a differentiable quantizer designed around a parametric sum of hyperbolic tangents, called \stanh, that relaxes the step-wise quantization function.
\stanh is implemented as a differentiable activation layer with learnable quantization parameters that can be plugged into a pre-trained fixed rate model and refined to achieve different target bitrates.
Experimental results show that our method enables variable rate coding with comparable efficiency to the state-of-the-art, yet with significant savings in terms of ease of deployment, training time, and storage costs.
\end{abstract}


\input{chapters/1_introduction/1_introduction}
\input{chapters/2_related_works/2_related_works}

\input{chapters/3_method/3_method}
\input{chapters/4_experiments/0_experiment_setup}
\input{chapters/4_experiments/1_choosing_anchors}

\input{chapters/4_experiments/2_RD_performance}

\input{chapters/4_experiments/3_training_cost}

\input{chapters/4_experiments/4_memory_usage}

\input{chapters/4_experiments/5_comprehensive_result_spiders}

\input{chapters/5_conclusion/5_conclusion}

\input{chapters/6_bibliography/6_bibliography}
\vfill

\end{document}

%% file: chapters/1_introduction/1_introduction.tex
\section{Introduction}

\IEEEPARstart{L}{earned} image compression (LIC) has seen much interest since it has achieved compression efficiency comparable to standardized codecs~\cite{review}.
At the transmitter side, the image is first projected into a lower-dimensional latent representation via a convolutional encoder.
Next, the latent representation is quantized and entropy-coded, yielding a compressed representation of the picture in the form of a binary bitstream.
At the receiver side, this representation is reversed and projected back to the pixel domain by a decoder, obtaining a lossy representation of the original image.
Such encoder-decoder (\textit{autoencoder}) models are trained end-to-end via back-propagation of the error gradient to minimize some \emph{rate-distorsion} (RD) cost function in the form 
\begin{equation}
    \mathcal{L} = R + \lambda \cdot D,
    \label{eqn:RD}
\end{equation}
\noindent
where
$\lambda$ is the hyper-parameter that regulates the trade-off between rate $R$ of the latent representation and distortion $D$ of the reconstructed image.
For example, a larger $\lambda$ places more importance on reconstruction quality at the expense of the rate of the latent space.
\\
The quantization of the latent representation is a crucial aspect: since quantization represents a non-differentiable function, it cannot be easily incorporated into error gradient back-propagation. 
A common approach in LIC is to replace quantization with additive uniform noise, ensuring resilience to quantization errors. 
Most existing approaches to learned image compression all share the same shortcomings, 
i.e. a separate encoder-decoder model with millions of parameters must be trained for each different bitrate, with two important implications.
First, training a different model from scratch for each rate incurs high costs in terms of both energy and time.
Second, the need to store a different model for each rate is a significant drawback, especially for resource-constrained devices. 
These issues jeopardize the feasibility of learned image compression in real-world scenarios, where controlling the rate is of paramount importance.
While solutions have been proposed recently, the issue of variable bitrate (VBR) image compression is far from being solved, prompting the present research.  
\\
In order to build a variable rate model, this work exploits \stanh, a differentiable quantization layer that can be plugged into a pre-trained codec to achieve variable bitrates.
\stanh is designed around a finite summation of hyperbolic tangents that relaxes the quantization function during training.
The relaxation is controlled by a single \emph{temperature} that is annealed during training to approach the desired quantization levels.
\stanh can be implemented as a differentiable quantization layer with just a few hundred learnable parameters, allowing standard training via end-to-end error back-propagation.
With respect to comparable methods, \stanh directly manipulates the latent representation, determining uniquely the quantization levels.
This allows moving from fine to coarse-grained quantization simply refining the parameters of the quantization layer, without retraining the other model parameters.
Our method allows switching encoding rate by simply switching the quantization layer, reducing the memory requirements for storing models, avoiding long training times, and reducing energy requirements.
We experiment with three different state-of-the-art image compression architectures and we show that our method allows variable rate coding with negligible impact on encoding efficiency.
\\
The rest of the paper is organized as follows. In Sec.~\ref{related} we provide the required background on learned image compression and on latent representation quantization.
In Sec.~\ref{method} we describe \textit{STanH}, demonstrating its properties and explaining how to plug it into a LIC model and how to exploit this module to adapt a pre-trained model for variable rates. In Sec.~\ref{experiments} we present quantitative results applying our method to multiple state-of-the-art fixed- and variable-rate models. Finally, in Sec.~\ref{conclusion} we draw the conclusion and discuss future works.

%% file: chapters/2_related_works/2_related_works.tex
\section{Background and Related Works} \label{related}

In this section, we first provide the relevant background on LIC. 
Then, we overview the existing approaches to latent representation quantization.
Next, we take a look at recent approaches toward variable rate image compression and highlight the limitations of the state-of-the-art that stimulated our research.

\subsection{Learned image compression fundamentals} \label{LIC}

Learnable Image Compression models have shown the potential to match or even outperform standardized codecs such as the recent H.266/VVC~\cite{vvc} in RD terms.
Early seminal works such as~\cite{balle2017,Theis2017} exploited a simple convolutional autoencoder structure with a unique latent representation modeled with a fully factorized distribution among channels modeled either analytically~\cite{SFC} or through an auxiliary neural network, and exploiting Generalized divisive normalization (GDN)\cite{GDN} activation functions. 
The scheme was improved by~\cite{balle2018} introducing a pyramid-based architecture composed of two nested variational autoencoders. Here, the first is called \emph{hyperprior} and captures spatial correlation within the image, while the second one models the latent representation, which is supposed to follow a zero-mean Gaussian distribution.
\cite{minnen2018,Lee2019,MinnenSingh2020}~combined previous architecture with a context-based auto-regressive entropy model to capture more local spatial correlation by exploiting already decoded parts of images. 
 Similarly, \cite{Li2020}~improved context modeling through a 3D zigzag scanning order, and improved parallelism entropy decoding with a 3D code dividing technique by partitioning the latent representation in multiple independent groups. In~\cite{li2021} a special non-local operation is proposed to consider global similarity within the context, by introducing U-net-like blocks, while~\cite{chen2021} introduced non-local network operations as non-linear transforms in both latent representations.
In more recent works other techniques have been exploited to improve RD performance; \cite{cheng}~replaced simple Gaussian distribution with a mixture of Gaussians and introduced some attention module to enhance entropy estimation, while~\cite{fu2023} introduced a more flexible discretized Gaussian-Laplacian-Logistic mixture model for the latent representation. 
In~\cite{zou} local attention is exploited to combine the local-aware attention with the global-related feature learning and to the present date is among the best-performing architectures.
In the following, we chose this architecture to exemplify how we integrated our quantization method into a typical image compression architecture.
Yet, we will demonstrate the versatility of our method by applying it to other two architectures in the experimental section.
\\
Tang~et~al.~\cite{Tang23} proposed a self-attention mechanism based on graphs to improve entropy estimation, and other threads of works tried to enhance some aspects of previous models: \cite{xie2021}~for example extracted better transformation between image and latent features space by exploiting invertible networks, while~\cite{RevNet,Swin2022} replaced convolutional modules with the Swin~transformer to achieve better compression efficiency with fewer parameters. 
On the other hand, \cite{Wu2022}~focused on the optimization of the image decoding through a learned block-based framework, \cite{wang22}~tried to adapt the entire structure to single images with an additional model stream to generate the transform parameters at the decoder side, and~\cite{he2021} introduced a checkerboard context model to improve efficiency during the decoding stage.
In~\cite{duan2023} hierarchical VAE architecture, originally designed for generative image modeling, is exploited for LIC, redefining their probabilistic model to allow easy quantization and practical entropy coding. 
In addition to the aforementioned variational autoencoder-based approaches, other approaches have been explored for compressing images using neural networks: \cite{Agustsson2019,Mentzer2020}~exploited GAN-based architecture to reduce image compression artifacts, especially at extremely low bitrates.
\\
The common feature shared by all the mentioned works is that a single model targets a single rate, meaning that multiple models must be trained and stored to cover the range of rates required in practical image or video compression scenarios.
This issue is critical since each model includes tens of millions of learnable parameters.
The issue with variable rate image compression can be also related to the way quantization takes place, as discussed in the following.

\subsection{Quantized latent representations for end-to-end learning}\label{rel-qm}

In end-to-end image compression, latent space quantization is critical because of its non-differentiability: the gradient of the quantization function is zero everywhere apart from the boundaries among quantization levels, where it is undefined. 
In this section, we review the main approaches to this problem, which revolve around replacing quantization with an approximated function.
\\
\subsubsection{Straight-through estimator} utilized in~\cite{Theis2017}, this method involves substituting the derivative of the quantization with a smooth approximation during the backward step while retaining the original function during the forward step.
 In particular, they exploited the linear function as a derivative, since it is easy to implement as it brings no modification to the gradient.
This method lacks elasticity since it does not allow for gradual relaxation during training,  thus enforcing quantization on integers even during training.
\\
\subsubsection{Additive Uniform noise} introduced in~\cite{balle2017}, consists of replacing the actual quantizer with additive uniform noise during training. 
The benefit of this method is that the density function associated with the noisy latent space represents a continuous relaxation of the discrete density mass found in the quantized space: moreover, independent uniform noise is commonly employed as a representation of quantization error due to its ability to approximate the marginal moments of the error~\cite{quantization}.
In addition to these two aspects, adding uniform noise allows us to reframe the training as a variational optimization problem~\cite{balle2017}, resulting in more effective learning of flexible latent space. 
All these benefits have made this method one of the most widely used for approximating quantization during learning~\cite{balle2017,balle2018,cheng,minnen2018}, becoming a de-facto standard. 
However, this method has one significant drawback: it allows quantization only on integers, without the possibility of adapting quantization intervals. This happens because there is no way to control the latent space since it is not parameterized, which forces training a different model for each RD trade-off.
\\
\subsubsection{Soft to Hard annealing} These methods are based on annealing a parameter, called temperature, to approximate quantization.
The main concept is to decrease this hyper-parameter during training to gradually constrain the latent representation, moving towards a hard quantization shape to ultimately freeze it.
For example, Using the softmax function over a partition of the latent space in Voronoi tessellation over centers enables soft quantization in~\cite{sth2017,sth2021}.
In~\cite{Agustsson2020}, they combine additive uniform noise and a new variant of softmax quantization to bridge the gap between quantized and continuous latent space, achieving robustness to quantization errors
Despite efficacy, this method suffers from several limitations; first, it is not agnostic to the entropy estimation since it cannot handle Gaussian distribution as prior, second, it introduces a further term in the loss function, not required by our method, and third it is not focused on variable rate adaptation. 
\\
The approach in \cite{mimo} deals with multiple-input multiple-output (MIMO) communications and is similar to ours in spirit as it relies on a finite summation of hyperbolic tangents
to overcome the non-differentiability of the quantization step.
While in this work we share the same conceptual framework, we deal with the specific and different challenges of image compression.
To the best of our knowledge, this work is the first to show the applicability of such a framework to image compression and in particular to show that a trained LIC model can be turned into a variable rate one simply by plugging a learnable quantization layer.
Also, the reference requires annealing a specific temperature for each hyperbolic tangent, requiring tuning as many parameters as the quantization steps. Conversely, our work requires tuning a single value to control this aspect, as detailed in Sect.~\ref{annealing}.

\subsection{Towards variable rate image compression}
\label{rel-ra}

Almost all of the above approaches to quantization entail the same drawback, i.e., a different encoder-decoder model must be trained for each rate.
Several proposals have been made towards variable rate learned image compression, and here we review some.
\\
In~\cite{Theis2017}, the latent representation of a single autoencoder is scaled before quantization by adding a learnable scaling parameter, one for each channel; this allows adaptation of the latent space based on the required bitrate. 
However, using a single value for each channel to adapt the quality level may lead to a reduction in R-D performance. 
\\
In~\cite{Choi2019}, they proposed an autoencoder that is conditioned on the Lagrangian multiplier $\lambda$, that is not treated as a regular hyper-parameter as usual but is instead used as input to the network to produce specific latent representations.
Additionally, the network was trained using mixed quantization bin sizes, enabling it to adapt the rate by adjusting the bin size of the quantization applied to the latent representation. 
This model increases the complexity of the optimization since it introduces both $\lambda$ and the bin size as inputs to determine the target bitrates. 
\\
Another approach is to deploy a model that considers different resolutions at the same time in the same training phase, making it adaptable to different RD trade-offs: \cite{Yang2020}~presented a problem of optimizing variable RD, which involves adding a modulated framework to the deep image compression structure. This framework enables the structure to adapt to various levels of compression. 
In~\cite{Cai2019}, the autoencoder is trained to break down the input image into multiple levels of representations, aiming to optimize the rate-distortion performance across all scales. 
However, both~\cite{Cai2019} and~\cite{Yang2020} made the training phase more complicated and unstable.
\\
In \cite{gain21}, a pair of parametric gain units are inserted before and after the quantization step to achieve discrete rate adaptation with one single model; by using exponential interpolation, continuous rate adaptation is achieved without compromising performance. In particular, for each target level, there is a specific pair of gain units multiplying the latent representation element-wise before and after quantization.
However, Beyond gain units, \cite{gain21}~modifies the entropy model, passing from symmetric to asymmetric Gaussian distribution, which makes it complex to assess the benefits of gain units in isolation.
\\
Following a different approach, In~\cite{Jooyoung2022}, they introduced a 3D importance map to achieve essential representations for compression at various quality levels, encoding thus only a partial version of the latent representation based on the desired quality level. Furthermore they exploited quality adapter quantization by multiplying the latent representation by a quantization vector, similarly to \cite{gain21}.
Despite effectiveness, \cite{Jooyoung2022} does require learning the importance map, which involves training thousands of parameters.
Similarly to \cite{gain21}, \cite{icassp2020} embeds a set of quality scaling factors (SFs) into a model, by which they can encode images across an entire bitrate range with a single model; however, this approach shares the same limitations of \cite{gain21}.
In \cite{gao2022}, a ``per image'' optimal representation is obtained by applying SGD to the latent space and determining the quantization step using grid-search. This approach is limited by the need for optimization when coding every single image, making it unfeasible in most use cases that impose constraints on computational costs or real-time capacity.
\\
In \cite{evc2023}, they exploited mask parameter decay, adjustable quantization step, and knowledge distillation to train a smaller model from a teacher model; variable bit-rate is obtained using adjustable quantization steps. 
A large model is pruned using learnable mask decay layers.
However, this method relies on a multi-stage training phase that is more complex than ours, and it is not agnostic with respect to the architecture, jeopardizing the possibility to add this technique in a general image compression model.
\\
In general, the problem of adapting a single model to different rates has received less attention as more efforts have been focused on improving compression performance. 
Nonetheless, the ability to efficiently switch among different rates is a fundamental requirement for practical image coding.
In the following, we introduce \stanh, our learnable quantization scheme that we exploit towards variable rate image compression.

%% file: chapters/3_method/3_method.tex
\begin{figure*}[t!]
  \centering
 \includegraphics[scale = 0.28]{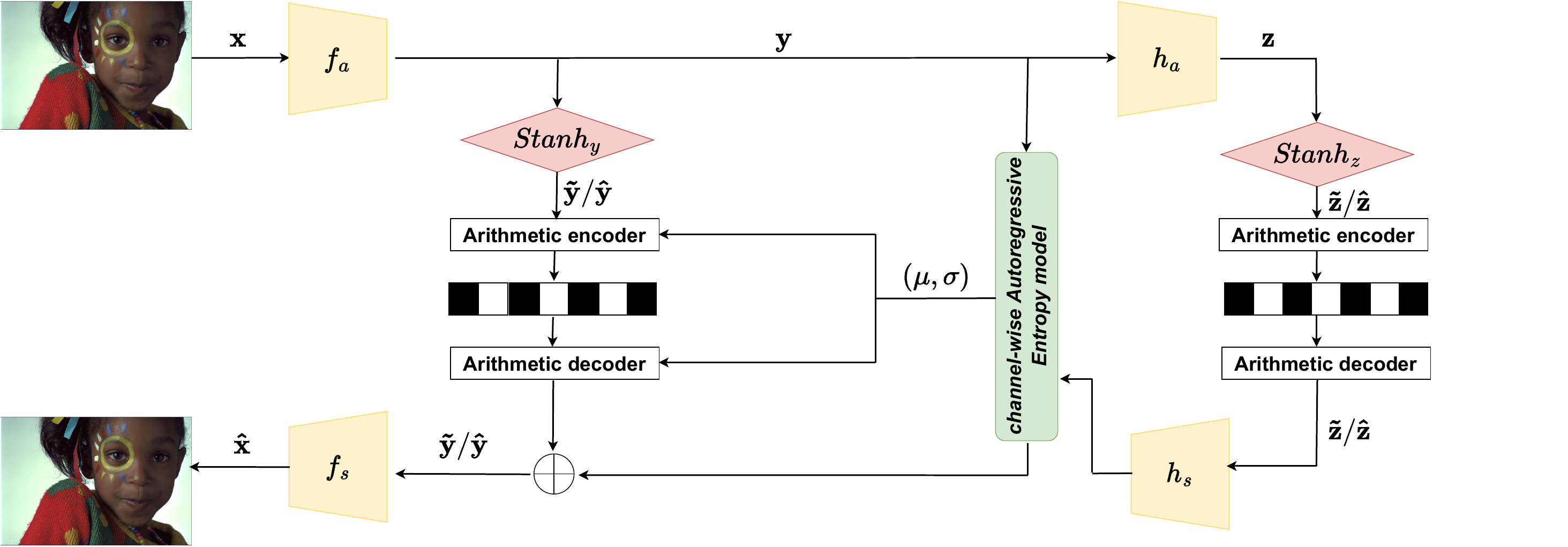}
  \caption{The reference learned image compression architecture \emph{Zou22} \cite{zou} (CNN-based architecture) with two \stanh layers for quantizing the main latent space $\textbf{y}$ and the hyperprior latent space $\textbf{z}$.}
  \label{fig:1}
\end{figure*}

\begin{table}[t]
  \caption{Overview of the notation used in this work.  }
  \begin{center}
    
    \label{tab:table1}
    \begin{tabularx}{\columnwidth}{r X} 
    \toprule
      \textbf{Symbol} & \textbf{Meaning}\\
      \midrule
      $\mathbf{x}/\hat{\mathbf{x}}  $ & input/decoded image.   \\
      $\mathbf{z}/\hat{\mathbf{z}}/ \tilde{\mathbf{z}}$ & original/discrete/soft-quantized hyperprior latent representation.  \\
    $ p_{\hat{\mathbf{z}}} / p_{\tilde{\mathbf{z}}}$ & discrete/relaxed hyperprior entropy model. \\ 
      $\mathbf{y}/\hat{\mathbf{y}}/ \tilde{\mathbf{y}}$ & original/discrete/soft-quantized main latent representation.  \\
      $ p_{\hat{\mathbf{y}}} / p_{\tilde{\mathbf{y}}}$ & discrete/relaxed image-reconstruction entropy model. \\ 
      $f_a / f_s$ & image reconstruction encoder/decoder. \\
      $h_a / h_s$ & hyperprior encoder/decoder. \\ 
      $N$ & dimension of the hyperprior latent representation. \\
      $M$ & dimension of the Gaussian-based latent representation. \\
      $r$ & residual vector. \\  
      $t$ & t-th training step. \\ 
      $L$ & number of quantization levels. \\ 
      $\mathbf{w} / \mathbf{b}$
      &  trainable weights / biases of  \stanh .\\
      $l_i$ & $i$-th trainable quantization level of \stanh .\\
      $\beta$ &  inverse temperature hyper-parameter of \stanh .\\ 
      $r_{i}^{-} / r_{i}^{+}$ & left/right bounds of quantization interval related to   $\tilde{y}_i$. \\
      $A_{i}$ & $i$-th anchor. \\
      $D_{ij}$ & $j$-th derivation from the $i$-th anchor. \\
      \bottomrule
    \end{tabularx}
    
  \end{center}
\end{table}

\section{Proposed method} \label{method}

In this section, we first present a reference image compression architecture and introduce the necessary notation.
Then, we mathematically define the quantization function \stanh and we introduce the \textit{annealing} strategy that allows it to approximate a scalar quantizer.
Then, we show how to plug \stanh into a learned image compression model as a latent space quantizer and how to train the model end-to-end for a govern RD trade-off.
Finally, we show how \stanh can be replaced by a simpler quantizer at inference time once the model has been trained.

\subsection{Preliminaries and notation} \label{prel}

In this section, we detail \emph{Zou22}~\cite{zou} (CNN-based architecture), the learned image compression architecture that we briefly introduced in the previous section and that we use as a reference since it represents the state of the art in learned image compression.
Fig.~\ref{fig:1} shows \emph{Zou22} integrated with our \stanh quantizer whereas Tab.~\ref{tab:table1} summarizes the notation used in the rest of this work.
The encoder $f_a$ projects the image $\mathbf{x}$ onto a low dimensional latent representation $\mathbf{y}$ of dimension $M$, which is then quantized, obtaining $\hat{\mathbf{y}}$: it is referred to as \emph{main latent representation} since the image is directly reconstructed from it.
Moreover, $\mathbf{y}$ is further projected into a second latent representation $\mathbf{z} = h_{a}(\mathbf{y})$ of dimension $N$, which is then quantized to $\mathbf{\hat{z}}$.
This latter \textit{hyperprior} latent space is exploited to find the spatial correlation for entropy estimation~\cite{MinnenSingh2020}, where channel-conditioning and latent residual prediction have been introduced, for enhancing rate approximation and reducing quantization error, respectively.
Towards this end, the feature map $\mathbf{\hat{y}}$ is divided into a predetermined number of slices, and then the spatial context of a specific slice is extracted by integrating information from both the hyperprior and a channel context model that receives previously decoded slices as input.
In addition, also the residual $r$ obtained during the quantization step is estimated, which is then added to the latent representation to reduce quantization error.
The output of this entropy model is represented in terms of means $\mu$ and standard deviation $\sigma$ representing spatial correlation for each element of $\mathbf{\hat{y}}$ since the latter is modeled as a Gaussian distribution. 
Finally, $\hat{\mathbf{y}}$ is fed to the decoder (\emph{synthesis transform}) obtaining the reconstructed images $\hat{\mathbf{x}} = g_{s}(\hat{\mathbf{y}})$.
From now on, we indicate with $N$ and $M$ the dimensions of the hyperprior and the main latent representation, respectively.

\begin{figure}[h!]
  \centering
 \includegraphics[width=0.6\columnwidth]{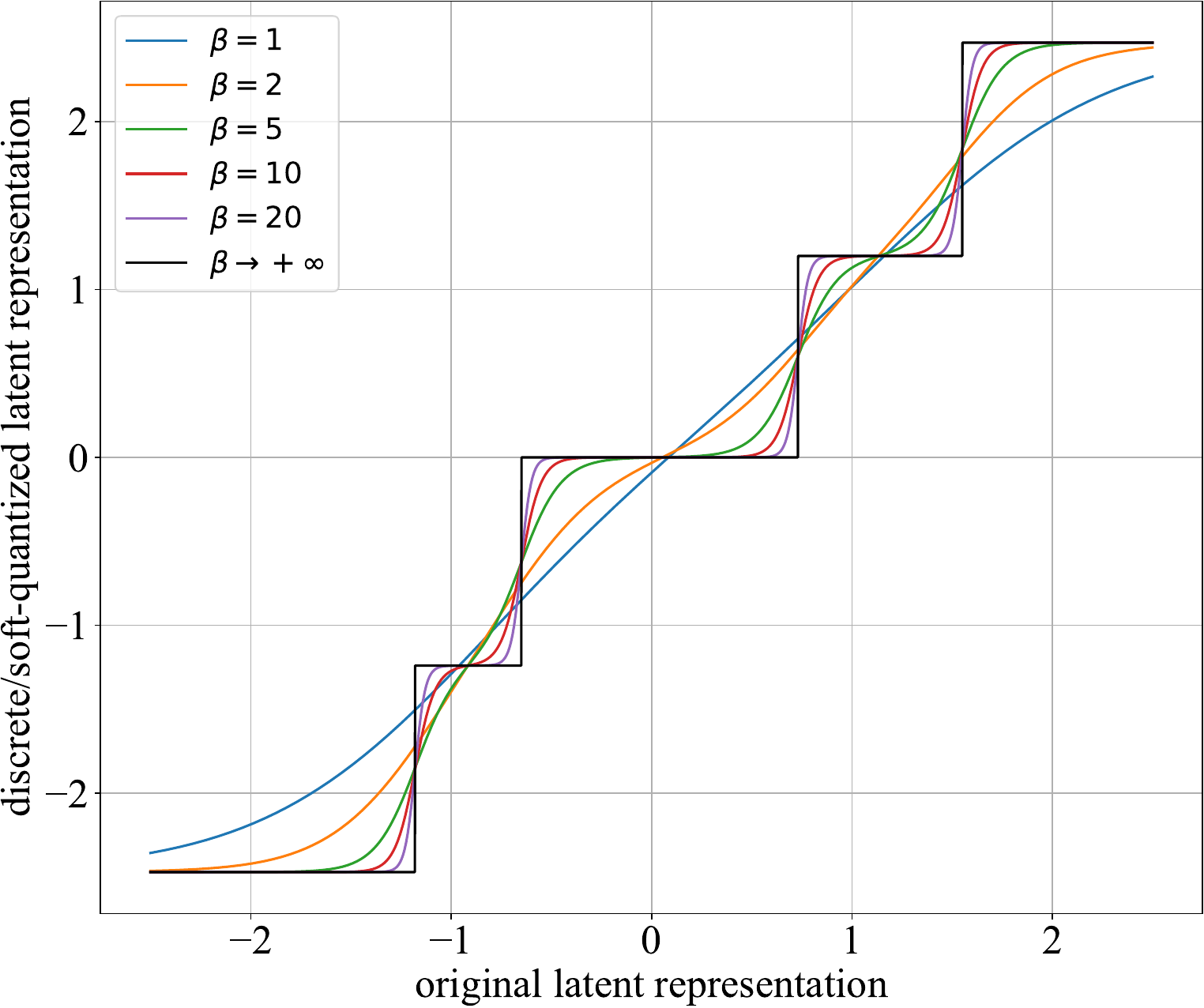}
  \caption{ \stanh activation function with $L=5$ quantization levels and for increasing values of inverse temperature $\beta$.}
  \label{fig:2}
\end{figure}

\subsection{Sum of Hyperbolic tangents for differentiable quantization} \label{stanh}

Our goal is to exploit a scalar quantization function that is differentiable and allows backpropagating the gradient of the error function at training time when plugged into an image compression architecture such as \emph{Zou22}. 
Let $\mathbf{y} \in\mathbb{R}^{C \times H \times W }$ be the tensor representing the real-valued latent space, where $C$, $H$, and $W$ represent the channels, the height, and the width dimension respectively.
In this context, we can exemplify our goal
as relaxing the discrete latent representation $\mathbf{\hat{y}}$ through a continuous proxy (soft-quantized) $\mathbf{\tilde{y}}$  at training time.
Toward this goal, the desired quantization function must satisfy the following requirements:
\begin{enumerate}[i)]
    \item the slope of the quantization steps shall be controllable at training time;
    \item the width of the quantization interval and the corresponding reconstruction level shall be parametric and learnable at training time.
\end{enumerate}
\noindent
Requisite i) is instrumental in making the quantizer arbitrarily close to an actual scalar quantizer, i.e. ladder-like function.
Requisite ii) allows our model to adapt the quantization intervals during training, thus replacing the conventional technique of rounding to the integer.
This aspect is fundamental for our method since it allows us to adapt the same model to different rates, as we experimentally show later. 
\\
\textbf{Relaxed quantization at training time.}
In order to meet the two requirements, we design the parameterized activation function \stanh illustrated in Fig.~\ref{fig:2}.
Let $L$ be the number of desired quantization levels, then the \stanh quantization function is defined  as the summation of $L\!-\!1$ translated and weighted hyperbolic tangents and is applied to $\mathbf{y}$ element-wise as follows:


\begin{equation} \label{stanh_relaxed}
 \tilde{\mathbf{y}} = \Stanh(\mathbf{y}, \beta)  
=   \sum_{i = 1}^{L-1}\frac{w_{i}}{2} \cdot \tanh[\beta \left( \mathbf{y} - b_{i}\right)]. 
\end{equation}
\noindent
About requisite i) $\beta$, which is referred to as \emph{inverse temperature}, regulates the slope of the quantization steps, i.e. the relaxation of the discretized latent representation: the higher $\beta$, the closer the slope to that of the step-wise function.
For this reason, its value is gradually increased during the training with the \emph{annealing} procedure detailed in Sec.~\ref{annealing}.
About requisite ii) the parameters  $\mathbf{w}= (w_1, w_2, ..., w_{L\!-\!1})$ $\mathbf{b}= (b_1, b_2, ..., b_{L - 1})$ determine the reconstruction levels for $\mathbf{\tilde{y}}$  and width of the quantization intervals, respectively. 
In fact, assuming $\beta\!=\!\infty$ in \eqref{stanh_relaxed}, we have that the first reconstruction level is equal to
\begin{equation}
 l_{1} =  \Stanh(y,  \infty)_{\left| y < b_{1} \right.}   = - \frac{1}{2} \sum_{i = 1}^{L - 1} w_i .
\end{equation} 
where the subscript $y < b_{1}$ represents the application of \stanh to a general value smaller than $b_{1}$.
The $i$-th reconstruction level $l_{i}$ is then obtained by adding  $w_{i-1}$ to the previous level $l_{i-1}$  as follows:
\begin{equation}
l_{i} = l_{i - 1}  +  w_{i - 1} \quad \forall i = 2,...,L .
\label{quantization_level_after}
\end{equation}
Consequently, at inference time, with $\beta = +\infty$, \eqref{stanh_relaxed} is replaced by the actual scalar quantizer with reconstruction levels as in~\eqref{quantization_level_after} and quantization intervals determined by $\mathbf{b}$, obtaining thus $\mathbf{\hat{y}}$.
During training, with $\beta < +\infty$, transitions between reconstruction levels are exponentially smoothed as in Fig.~\ref{fig:2}, that exemplifies \stanh with $L=5$: for $\beta$ equal to one, the shape of \stanh is close to the hyperbolic tangent; as $\beta$ increases, the shape progressively approaches the step-wise quantization function.
Since \stanh has a derivative all over its domain, $\mathbf{w}$ and $\mathbf{b}$ can be learned to minimize an arbitrary loss function when $STanH$ is plugged into the back-propagation procedure as described below.

\subsection{Temperature annealing procedure}\label{annealing}

In this section, we explain how we anneal the inverse temperature $\beta$ during training to achieve a final configuration that is consistent with the actual quantizer.
To achieve this goal, the procedure for annealing $\beta$ is crucial towards convergence: annealing too fast could lead the model to settle on a local minimum, whereas annealing too slowly could prevent being robust against quantization errors. 
Following~\cite{Dickstein2015}, the issue is tackled by incrementally increasing $\beta$,
relying on a function that considers both the number of training iterations and the difference between the relaxed and quantized latent space. In this way, it is possible to progressively approach the discrete latent configuration during training. 
\\
\textbf{Semi deterministic-based annealing.} Taking inspiration from~\cite{sth2017}, we propose a strategy where $\beta$ is increased in a semi-deterministic way.
In a nutshell, we progressively drive the possible values of the latent representation towards the discrete quantization levels, yet without overly constraining the configuration of the latter during training.
Let $t$ indicate the $t$-th step of the training procedure based on the back-propagation of the error gradient described in the following.
For a given batch of training samples, the quantization errors of the soft-quantized and discrete quantized latent representation are computed as:
\begin{equation} \label{relaxed_error}
        \tilde{e}_t = \| \tilde{\mathbf{y}} - \mathbf{y}\|^2,
\end{equation}
\begin{equation} \label{quantized_error}
        \hat{e}_t = \| \hat{\mathbf{y}} - \mathbf{y}\|^2.
\end{equation}
By taking the absolute difference between (\ref{relaxed_error}) and (\ref{quantized_error}) we obtain the error between the discrete and continuous representation at training step $t$:
\begin{equation}
    E_{t} = | \hat{e}_{t} - \tilde{e}_{t}|.
\end{equation}
The smaller this error, the closer the latent space to the actual discrete space to be encoded. After computing $E_{t}$, we update the values of the temperature:
\begin{equation} \label{beta_annealing}
\begin{split}
& \beta_{1}  = 1 \\
& \beta^{max}_{1}  = 1 \\
& \beta^{\text{max}}_{t} = \beta^{\text{max}}_{t - 1} + K \cdot  \text{$E_{t}$}  \quad \forall t > 1 \\ 
& \beta_{t} \sim Uniform\left(1, \beta^{\text{max}}_{t} \right) \quad \forall t > 1, \\
\end{split}
\end{equation}
where $\beta_{t}$ represents the value used at the $t$-th step, and $K$ is a factor that regulates the velocity with which we freeze the latent representation.
It is easy to understand that, as the training steps progress, the latent representation will become on average increasingly frozen towards the final true configuration, making the model robust to the quantization error.

\subsection{End-to-end learning with STanH}
\label{e2e_learning}

In this section, we first show how we implement our differentiable quantizer as a layer that can be plugged into a generic image compression model.
Next, we formulate the rate-distortion cost function to be minimized at training time and we detail the end-to-end training procedure.
While we exemplify the training procedure for \textit{Zou22}, it can be generalized to any architecture.

\subsubsection{STanH as a quantization layer}

\stanh can be implemented as a parametric layer that can be plugged at any arbitrary position in a neural model to implement quantization.
In the case of \textit{Zou22} in Fig.~\ref{fig:1}, both the hyperprior $\mathbf{z}$ and the main latent representation $\mathbf{y}$ require to be quantized, therefore two independent instances of \stanh layer are required. 
Every instance operates independently and undergoes separate training to acquire its own distinct quantization functions.
In fact, $\mathbf{z}$ and $\mathbf{y}$ have different semantic meanings: the former represents the hyperprior latent representation, while the latter represents the one from which the image is reconstructed. Therefore, it is likely that these two latent representations have distinct distributions with distinct optimal quantization levels.
Having two separate \stanh layers involves only a few hundred additional parameters to train, which is a negligible figure compared to the entire model.
\\
When plugging \stanh, we obtain the following latent representations:

\begin{align} 
\tilde{\mathbf{z}}_t &= \Stanh_{h} \left(  \mathbf{z}, \beta_{h,t} \right) \nonumber \\
\tilde{\mathbf{y}}_t &= \Stanh_{m} \left(  \mathbf{y}, \beta_{m,t} \right) \label{stanh_module_application_train}
\end{align}
\noindent
and
\begin{align} 
\hat{\mathbf{z}}_t &=    \Stanh_{h} \left(  \mathbf{z}, \infty \right) \nonumber\\
\hat{\mathbf{y}}_t &=   \Stanh_{m} \left(  \mathbf{y},  \infty \right),\label{stanh_module_application_test} 
\end{align}
\noindent

where $t$ indicates the $t$-th training step, while the subscripts \{\emph{h},\emph{m}\} refer to the two different latent representation, namely the hyperprior and the main one.
We highlight that since $\Stanh_{h}$ and $\Stanh_{m}$ are implemented as two independent layers, $\beta_{h,t}$ and  $\beta_{m,t}$ are also modified through annealing independently one from the other.


\subsubsection{Optimization problem formulation}
Plugging the \stanh layer in an image compression model preserves the nature of the standard rate-distortion (RD) optimization problem 
\begin{align} \label{loss_function}
 \mathcal{L} &=\lambda \cdot d(\mathbf{x}, \hat{\mathbf{x}})   
 + \mathcal{R}(\mathbf{\hat{z}})  + \mathcal{R}(\mathbf{\hat{y}|\hat{z}})   \notag \\ 
&= \lambda \cdot d(\mathbf{x}, \hat{\mathbf{x}})  - \mathbb{E}[\log_2  p_{\hat{\mathbf{z}}}(\hat{\mathbf{z}})]  - \mathbb{E}[\log_2 p_{\hat{\mathbf{y}}| \hat{\mathbf{z}}}(\hat{\mathbf{y}}| \hat{\mathbf{z}})] 
\end{align}
\noindent
where the first term is some distortion metric $d$, while the second and the third represent the rate contribution of the two latent representations. 
The hyper-parameter $\lambda$ is the Lagrangian multiplier that controls the trade-off between rate and distortion.
A bigger $\lambda$ puts more penalty on large distortions, whereas a smaller $\lambda$ puts more penalty on the rate of the latent representations.
In particular, for the second term we follow the standard approach proposed in~\cite{balle2018} to use an ad-hoc neural network to directly estimate the rate: in this way, following the same configuration as previous works, we impose a fully factorized distribution among channels as follows:

\begin{equation} \label{fullyfac}
   p_{\hat{\mathbf{z}}} (\hat{\mathbf{z}}, \mathbf{\psi}) = \prod_{j=1}^C p_{\hat{z}_{j}}(\hat{z}_{j},\psi_{j}) ,
\end{equation}
\noindent
where $\psi_{j}$ represents the learnable parameters.
\\
The third term represents the rate of a Gaussian-like latent space, therefore there is no need for a further neural network to determine the rate.
In fact, for a specific $y_i$, it is possible to evaluate its cumulative distribution at training time as: 

\begin{align} \label{gaussian_rate}
 \mathcal{R}(\tilde{y}_i) &= \int_{\tilde{y}_{i} - r_i^{-}}^{\tilde{y}_{i} + r_i^{+}} p_{\tilde{\mathbf{y}}| \tilde{\mathbf{z}}} (y) dy  \notag \\ 
&=  \frac{1}{\sqrt{2\pi}\sigma_i} \int_{\tilde{y}_{i} - r_i^{-}}^{\tilde{y}_{i} + r_i^{+}} \exp\left[-\frac{(t-\mu_i)^2}{2\sigma_i^2}\right] dt \notag \\
&= \Phi(\tilde{y}_{i} + r_i^{+}) - \Phi(\tilde{y}_{i} - r_i^{-})  ,
\end{align}
\noindent
where $\Phi$ is the $cdf$ of the Gaussian distribution with mean $\mu_i$ and standard deviation $\sigma_i$, while $r_i^{-}$ and $r_i^{+}$ are the left and the right bounds of the quantization intervals related to $\tilde{y}_i$, respectively.
Notice that, differently from other works that rely on uniform quantization over integers, where $r_i^{-}$ and $r_i^{+}$ are both equal to $0.5$, in our case these two values change depending on \stanh.

\subsubsection{Learning procedure} \label{training_procedure}
We recall that our goal is achieving variable rate image coding without training a separate model for each target rate as in \cite{cheng,zou,xie2021}.
Let us assume that at least one model has been preliminary trained for some target rate that we call \textit{anchor} model.
Since in our scheme, the quantizer is implemented as a learnable layer, different rate-distortion tradeoffs can be achieved by plugging a different \stanh layer in the anchor and refining the \stanh parameters for different $\lambda$ values.
Practically speaking, we plug a \stanh layer in the anchor, we freeze all the anchor layers but \stanh and we refine this latter layer for a few epochs.
We repeat the procedure for each different target rate refining a separate \stanh layer for a different $\lambda$ value for each target rate.
In detail, we refine the \stanh layer for increasingly lower $\lambda$ values in~\ref{loss_function} (i.e., we gradually reduce the target rate).
The result of this procedure is a set of refined models that share the same learnable parameters as the anchor except for the \stanh layer: we call these models \emph{derivations}.
Therefore, only one anchor model and one \stanh layer from each derivation need to be stored, once the refinement process is over.
Once the refinement procedure above is over, it is possible to switch encoding rates simply by plugging into the anchor model the \stanh layer from the correct derivative.
Notice that in principle, it is possible to consider more anchors trained for different rates, and in this case, it is desirable that the anchors are evenly spaced across the target rate range.
In the following, we experiment with different numbers of anchors and target rates with different LIC models.

\subsection{From fine-grained to continuous rate adaptation} \label{fctf}

In practical image coding applications, it is of paramount importance to control the rate at a fine granularity, e.g. by tuning the QP in standardized codecs~\cite{vvc}.
\stanh allows  rate adaptation in two ways.
The first, straightforward, way is refining a new \stanh layer for the appropriate $\lambda$, as described above, achieving fine-grained rate control.
When finetuning a novel derivation is not possible (e.g., lack of training resources or samples), it is possible to strike continuous rate control by interpolating two existing derivations as follows.
Let us suppose we have two derivations with corresponding layers \stanh$_{1}$ and \stanh$_{2}$, achieving two different RD point $(r_2, q_2)$ and $(r_1, q_1)$, with $q_1 > q_2$ and $r_1 > r_2$.
The goal here is to produce a third derivation \stanh$_{3}$ at the RD point $(r_3, q_3)$ with $r_1 > r_3 > r_2$ and $q_1 > q_3 > q_2$.
Let $(w_1,b_1)$ and $(w_2,b_2)$ be the learnable parameters related to \stanh$_{1}$ and \stanh$_{2}$, respectively.
Instead of refining the new \stanh$_{3}$, a new set of parameters $(w_3, b_3)$ can be interpolated as follows; from now one we call these type of model interpolations.

\begin{equation}
\begin{aligned} 
w_3 &=  (1 - \rho) \cdot w_1 + \rho \cdot w_2  \\
b_3 &=  (1 - \rho) \cdot b_1 + \rho \cdot b_2 
\label{fgs}
\end{aligned}
\end{equation}
\noindent 
where $\rho \in (0,1)$ is tuned to match the desired target rate $r_1 > r_3 > r_2$ (with correspondent qualities  $q_1 > q_3 > q_2$).

%% file: chapters/4_experiments/0_experiment_setup.tex
\section{Experiments} \label{experiments}

In this section, we experiment with \stanh over three recent LIC models Zou22~\cite{zou}, \emph{Xie21}~\cite{xie2021}, and \emph{Cheng20}~\cite{cheng} as implemented in the CompressAI\cite{compressai} project.
We measure the impact of \stanh both in terms of RD performance and complexity.
Finally, in Sec. \ref{sec:comp_vr_models} we compare our method with two other variable rate codecs, while in Sec. \ref{overview} we analyzed \stanh capability of moving from coarse to fine granularity.

While in a practical scenario one would simply refine a pre-trained anchor following the procedure above, in these experiments we retrain from scratch the anchors with the \stanh module for two reasons; first,  we want to measure the cost of training from scratch and to set an accurate baseline model for benchmarking RD efficiency. Second, we also want to evaluate \stanh module in a fixed rate context scenario in which we considered only anchor models without derivations as shown in Sec. \ref{number_anchors}.
\footnote{The source code will be made available upon paper acceptance. Additional results and material  can be accessed at\href{https://drive.google.com/drive/folders/18IkZvLhzFV8HUvNE9PYkeaaF4XDuGypQ?usp=drive_link}{https://drive.google.com/drive/reconstructions}}

\begin{table}[t]
\caption{Parameters used to define anchors for each reference model, with their respective values of $\lambda$.
For each cell, the first row corresponds to the tuple ($N$,$M$) introduced in Sec.~\ref{prel}, while the second one is the $\lambda$ used for training.}
\begin{center}
\begin{tabular}{C{1cm} C{1.3cm}C{1.3cm}C{1.3cm}}
 \toprule
 \bf \emph{Model} & $\mathbf{A_{1}}$& $\mathbf{A_{2}}$ & $\mathbf{A_{3}}$ \\
    \midrule
 \emph{Cheng20}  & (128,128)  $\lambda$ = 0.0036  &    (192,192)  $\lambda$ = 0.013   &   (192,192)  $\lambda$ = 0.0483\\ 
 \midrule
 \emph{Xie21} &  (128,128)  $\lambda$ = 0.0030 &  (128,128) $\lambda$ = 0.010 & (192,192) $\lambda$ = 0.045 \\ 
 \midrule
  \emph{Zou22} &  (192,320)  $\lambda$ = 0.0025 &  (192,320) $\lambda$ = 0.010 &   (192,320) $\lambda$ = 0.0483 \\ 
 \bottomrule
 \end{tabular}
 \label{tab:table2}
\end{center}
\end{table}

\subsection{Experimental setup}
\label{setup}

In this section, we describe how we trained the three architectures with \stanh as a quantizer.
From now on, we refer to  $i$-th anchor as $A_i$, where $A_1$ represents the point with the highest rate and quality in the RD plane (i.e., the top-right point on a curve).
We refer to the $j$-th derivation from the $i$-th anchor as $D_{ij}$, where a larger value of $j$ indicates a greater deviation on the RD plot from the reference anchor (i.e., towards the bottom-left corner of the RD plot).
Coherently with the existing literature, we optimized the MSE as distortion metric $d$ in \eqref{loss_function}, and we consider six different $\lambda$ values, i.e. points, on the RD plot. 

At inference time, we used arithmetic coding to encode latent representations, using the \texttt{torchac} library~\cite{torchac}.
For \emph{Cheng20} and \emph{Xie21} we fix $L$=60  quantization levels for both the latent representations, and we initialize $\mathbf{w}$  and $\mathbf{b}$ in order to have an initial uniform quantization in the range $[ -30, 30]$, for a total of $240$ additional parameters.

For \emph{Zou22} we increased $L$ to 120 for the main latent representation $\mathbf{\hat{y}}$ (Gaussian-distributed) and we reduced L to 40 for the hyperprior latent representation; in this case, the number of parameters included in the two \stanh modules increased to 320.
We empirically fix $K = 15$, which regulates the annealing velocity of $\beta$ (Sec. \ref{annealing}) doubling it when the cost function reaches a plateau for the first time, with a patience of 50 epochs.

\begin{table}
\begin{center}
\caption{Values of $\lambda$'s used for training the derivations. We list only $\lambda$'s used in the case of three anchors implemented.}
\begin{tabular}{C{0.7cm} C{5.8cm}}
 \toprule
   \bf \emph{Model} & \bf $\boldsymbol{\lambda}$'s for Derivations  \\ 
    \midrule

    \emph{Cheng20} & ($\lambda_{D_{11}}, \lambda_{D_{21}},\lambda_{D_{31}}$) = (0.018, 0.0067, 0.0018)  \\ \midrule
    \emph{Xie21}& ($\lambda_{D_{11}}, \lambda_{D_{21}},\lambda_{D_{31}}$) = (0.017, 0.0060, 0.0010)  \\ \midrule
       \emph{Zou22} & ($\lambda_{D_{11}}, \lambda_{D_{21}},\lambda_{D_{31}}$) = (0.018, 0.0067, 0.0012)  \\ \bottomrule

    \end{tabular}
    \label{refined_lambda}
\end{center}
\end{table}
\noindent
\textbf{Training the anchors.} 
We trained each anchor on 24k random samples from the OpenImages dataset\cite{OpenImages} for, depending on the architecture, $\sim1$-$1.5$M steps with a batch size of $16$ images using the Adam~\cite{adam} optimizer with an initial learning rate of $10^{-4}$ that is reduced by a 2 factor when a plateau is reached, with 50 epochs patience.
The number of anchors is a hyper-parameter that drives a trade-off between RD performance and training costs in terms of time and storage that is explored in detail in Sec.~\ref{number_anchors}. 
Tab.~\ref{tab:table2} lists the different $\lambda$ values for used for training anchors for each reference model, considering the case of three anchors per model (actual values are from the reference papers).
All models are trained on an NVIDIA~A40 GPU.
\\
\textbf{Refining the derivations.}
Finally, we refine the \stanh layers to target  different rates.   
As each \stanh layer is only a few hundred learnable parameters, we found only about 8000 samples from the training dataset are enough to refine the layer.
We refine each derivation for $~2$-$3$ K steps and reduce the patience for a learning rate reduction from 50 to 10 epochs.
In Tab.~\ref{refined_lambda} we listed for each architecture the values of $\lambda$ used for refining the derivations, considering the case of three anchors (one derivation for each of them). We experimentally observed that it is possible to refine a derivation starting from either a higher quality anchor or from the nearest anchor and moving in both directions with respect to the target bitrate, the latter approach yielding somewhat better RD efficiency.
\\
\textbf{Evaluation.} We evaluate the above models on the Kodak PhotoCD image dataset~\cite{kodak}, Clic Professional validation and test dataset~\cite{clic}, and the Tecnik dataset~\cite{Tecnick}.
The Kodak dataset comprises 24 uncompressed images with a resolution of 768$\times$512, Clic dataset consists of 60 images of varying and higher resolutions, while Tecnik includes 100 images with a resolution of 1200$\times$1200.
The image quality (i.e., distortion) is evaluated as peak signal-to-noise ratio (PSNR) and secondarily as multiscale structural similarity (MS-SSIM)~\cite{mssim}.
The rate of the compressed latent representations is measured in terms of bits per pixel (bpp) to account for the different image resolutions.
Such metrics are plotted as rate-distortion curves and
pairs of curves are compared in terms of Bjontegard~\cite{bjonte} metrics.
We recall that a negative BD-Rate (fewer bpps required for the same PSNR) or/and positive BD-PSNR (higher PSNR for the same bpps) indicate better encoding efficiency.

%% file: chapters/4_experiments/1_choosing_anchors.tex
\vspace{-0.25cm}
\subsection{Experimenting with the number of anchors} \label{number_anchors}

\begin{figure*}[!b]
  \centering

  \begin{subfigure}{0.25\textwidth}
    \centering
    \includegraphics[width=\textwidth]{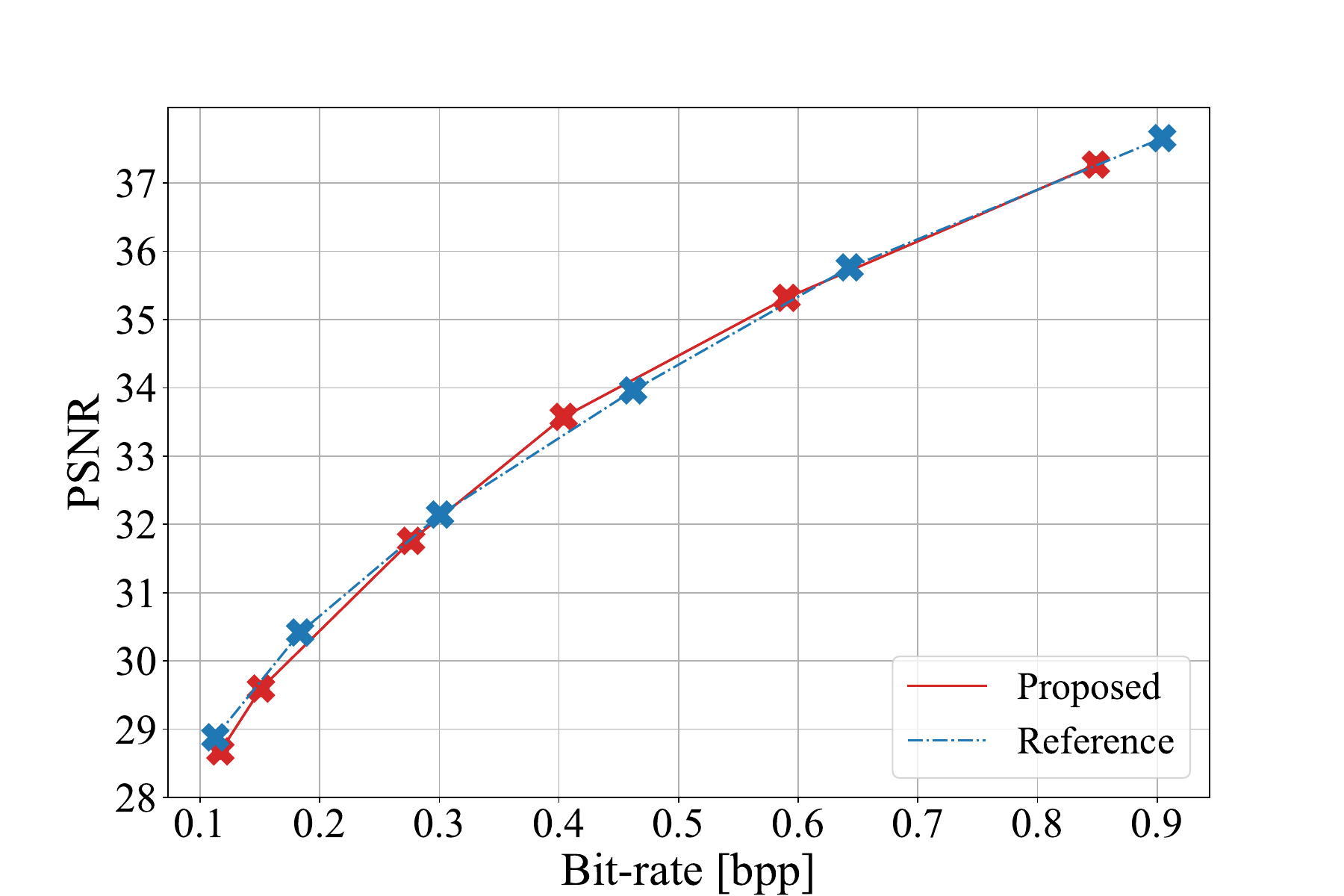}
    \caption{Six anchors.}
    \label{6a}
  \end{subfigure}
  \hfill
  \begin{subfigure}{0.25\textwidth}
    \centering
    \includegraphics[width=\textwidth]{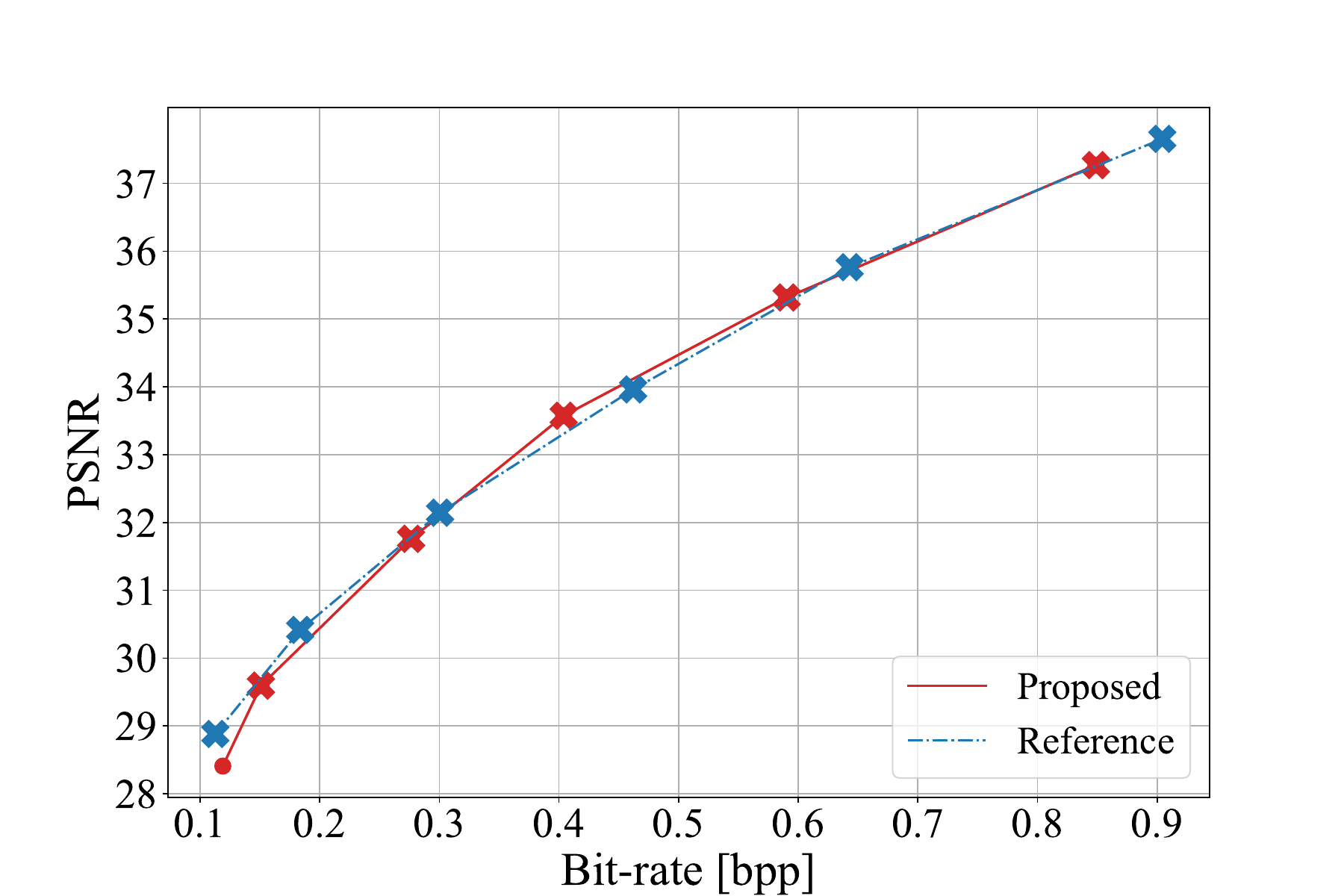}
    \caption{Five anchors.}
    \label{5a}
  \end{subfigure}
  \hfill
  \begin{subfigure}{0.25\textwidth}
    \centering
    \includegraphics[width=\textwidth]{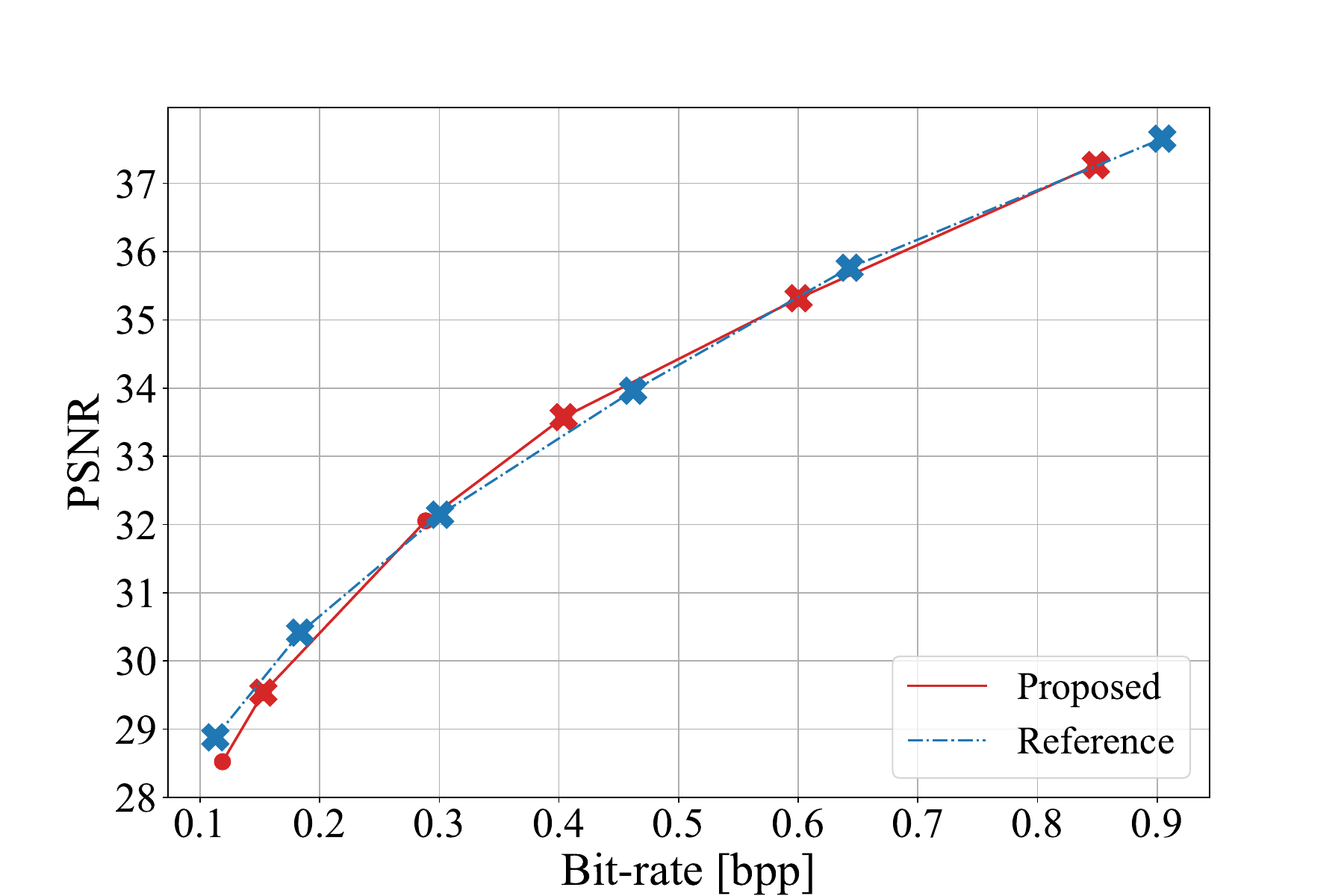}
    \caption{Four anchors.}
    \label{4a}
  \end{subfigure}

  \vspace{\baselineskip}

  \begin{subfigure}{0.27\textwidth}
    \centering
    \includegraphics[width=\textwidth]{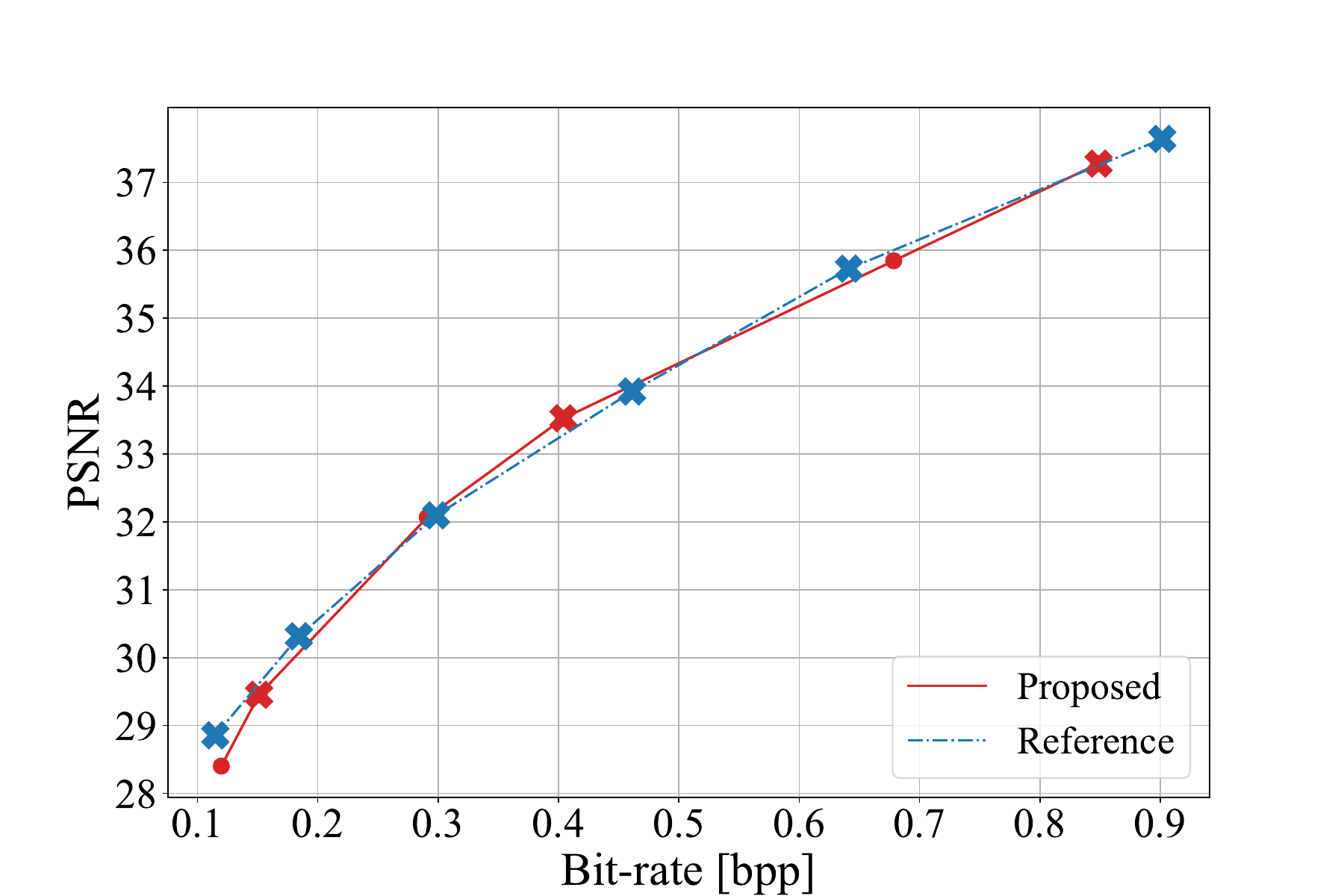}
    \caption{Three anchors.}
    \label{3a}
  \end{subfigure}
  \hfill
  \begin{subfigure}{0.27\textwidth}
    \centering
    \includegraphics[width=\textwidth]{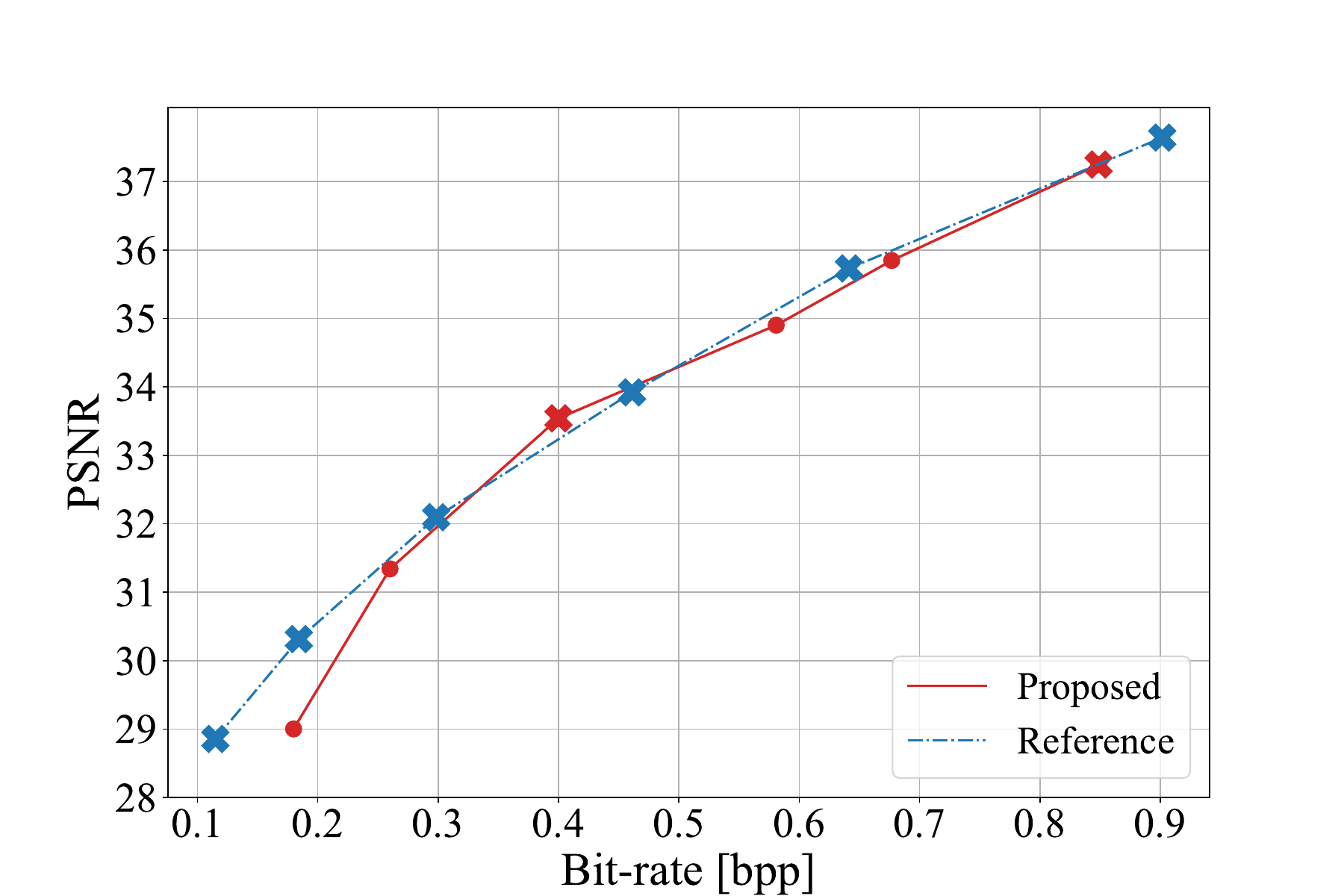}
    \caption{Two anchors.}
    \label{2a}
  \end{subfigure}
  \hfill
  \begin{subfigure}{0.27\textwidth}
    \centering
    \includegraphics[width=\textwidth]{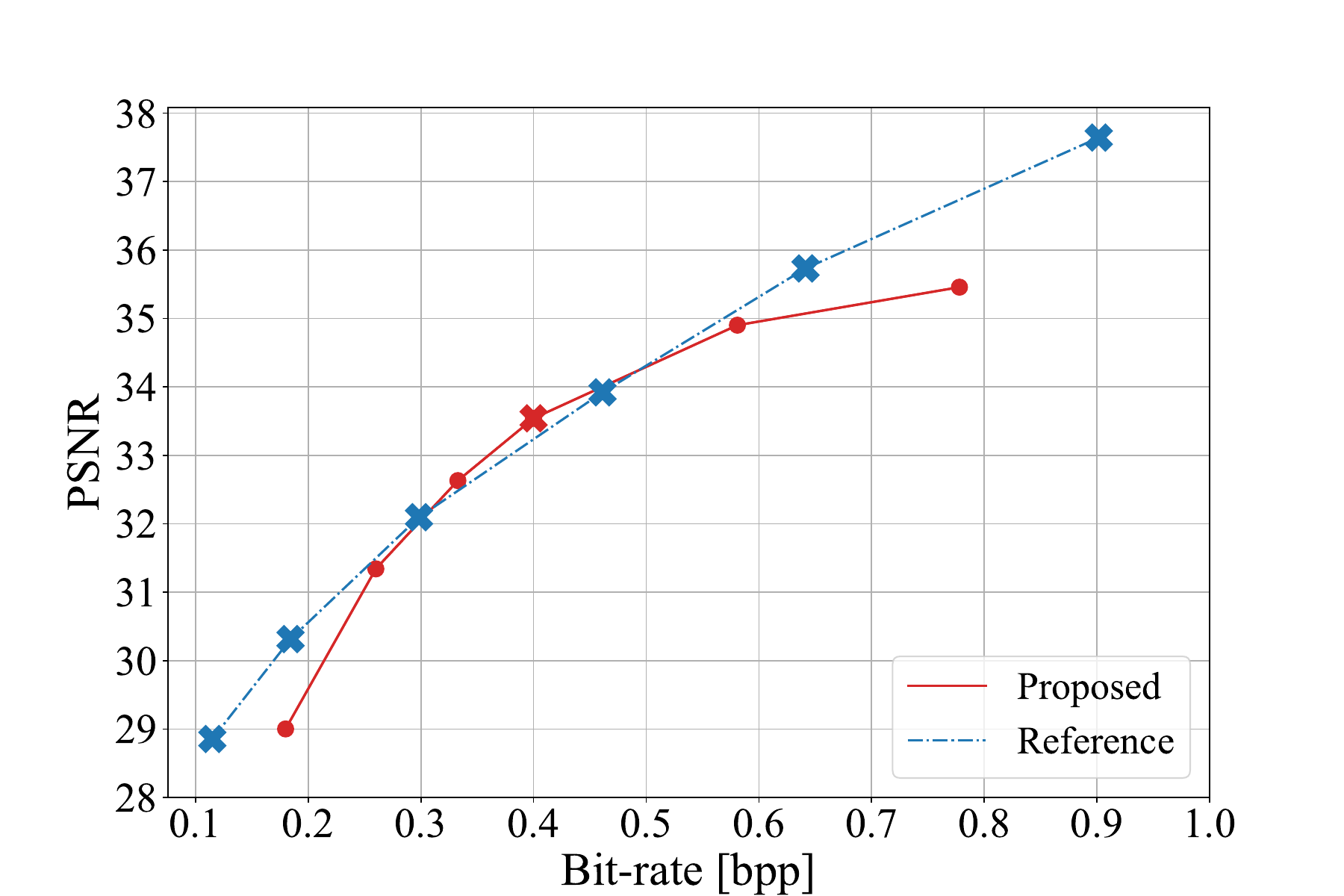}
    \caption{One anchor.}
    \label{1a}
  \end{subfigure}

  \caption{Rate-distortion performance of \emph{Zou22} on Kodak dataset using different number of anchors, from six (a) to one (f). For our proposed approach, red crosses represent trained anchors, whereas red circles the refined derivation(s).
  }
  \label{fig:3}
\end{figure*}

As a first experiment, we explore the performance-complexity tradeoff as a function of the number of anchors and derivations.
We experiment with decrementing the number of anchors from six (all models trained end-to-end, in a fixed-rate approach) to one (only one model trained end-to-end, other 5 are derived refining only the \stanh layers).
For the time being, we measure the complexity as the number of learnable parameters, since both overall training time and storage cost directly depend on that.
We take the \emph{Zou22} scheme as a reference, where the model is trained with different lambdas and without \stanh layer
\\ 
Fig.~\ref{fig:3} shows the RD performance of  for the reference scheme and the proposed \stanh-based scheme as a function of the number of anchors while Tab.~\ref{tab_anchors} shows the corresponding BD-Rate and BD-PSNR vs. $\emph{Zou22}$.
Anchor models are represented by a cross (blue for reference models, red for models with our module); only for the proposed scheme, derivations are depicted with a smaller circle.
The case with 6 anchors is reported as a sanity check to show that \stanh entails no RD penalty with respect to the reference fixed-rate quantization.
\\
As the anchor number decreases, the RD performance worsens slightly, yet the complexity drops much faster (see Tab.~\ref{tab_anchors}).
For 3 anchors, the BD-Rate drop is within 1\%, while the complexity is slashed by a factor of two.
\\
Finally, Fig.~\ref{1a} reports the corner case of one anchor only and 5 derivatives to stress the potential of our method.
As the derivations move away from the anchor, the RD performance of the derivations degrades proportionally to such distance, yet in a graceful manner.
\\
Since 
the three anchors setup (Fig. \ref{3a}) enables a reasonable performance-complexity trade-off (complexity reduced by $\sim 50 \%$ with a BD-Rate penalty below $1 \%$), we refer to this setup in the following.

\begin{table}
\centering
 \caption{BD-Rate and BD-PSNR vs. \emph{Zou22} on the Kodak test set for different numbers of anchors,  savings are reported in terms of Trainable Parameters (TP) for our method (proposed).
 }
\resizebox{\columnwidth}{!}{\begin{tabular}{cccccc}
  \toprule
  \textbf{\emph{Anchors}} & \textbf{\emph{Derivations}} & \textbf{\emph{BD-Rate}} & \textbf{\emph{BD-PSNR}} & \textbf{\emph{ TP-proposed}} & \textbf{\emph{Savings (\%)}} \\
  \midrule
  6& 0 &0.0026 & 0.0011 &  451.4 M  & 0\\
  5& 1 & 0.24 &  -0.0008& 376.2 M   &  16 \\
  4 & 2 & 0.43 & -0.05 &  300.9 M   &  33  \\
 3 & 3 & 0.99 & -0.07 & 225.7 M  & 50\\
  2 & 4 & 4.71& -0.12 & 150.6 M   &  67 \\
  1 & 5 & 9.09 & -0.28 &  75.2 M  & 86 \\
  \bottomrule
\end{tabular}}\label{tab_anchors}
\end{table}

%% file: chapters/4_experiments/2_RD_performance.tex
\begin{figure*}
      \begin{subfigure}{0.27\textwidth}
        \centering
        \includegraphics[width=\textwidth]{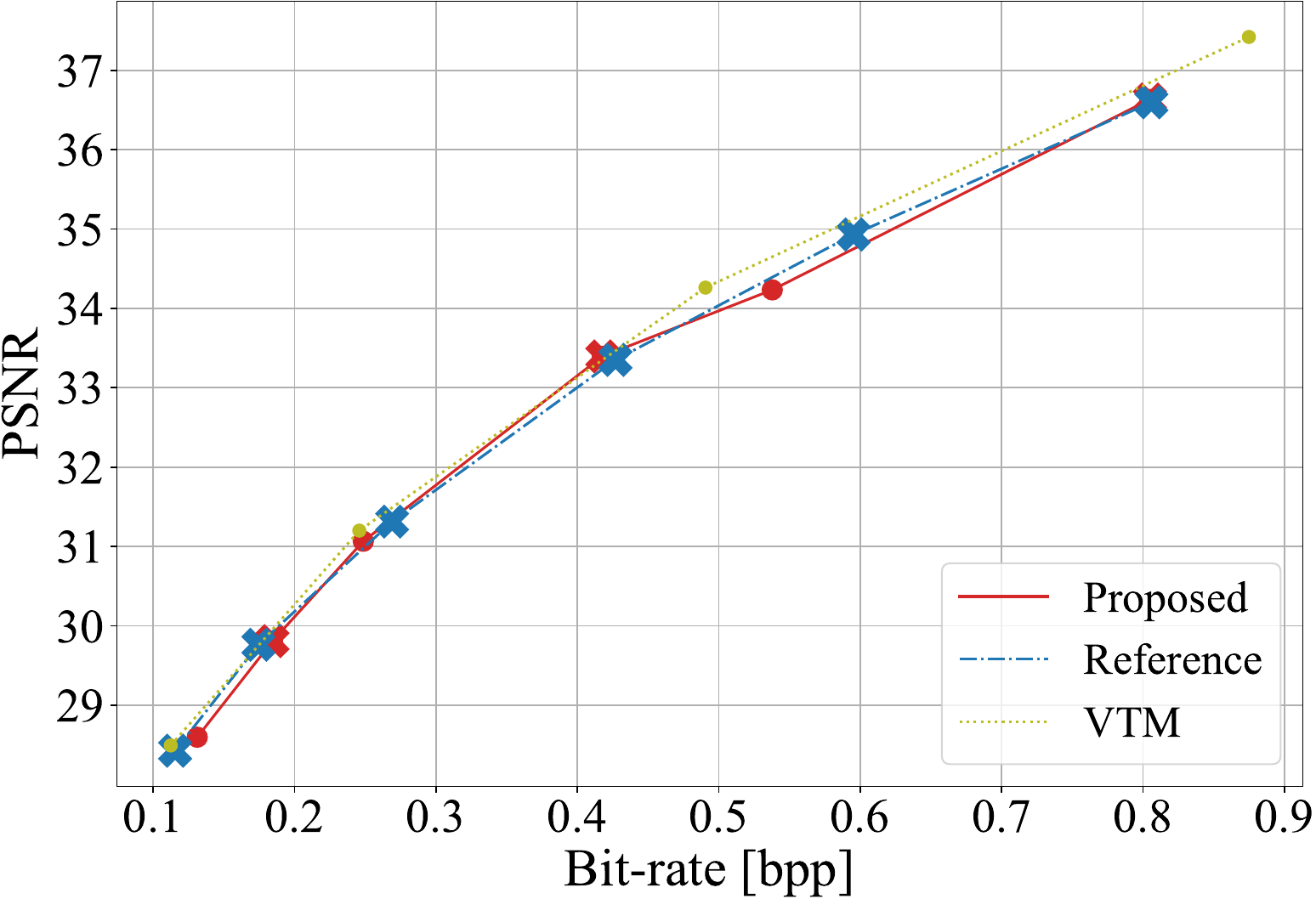}
        \caption{\emph{Cheng20}}
        \label{cheng2020psnrkodak}
  \end{subfigure}
  \hfill
    \begin{subfigure}{0.27\textwidth}
        \centering
        \includegraphics[width=\textwidth]{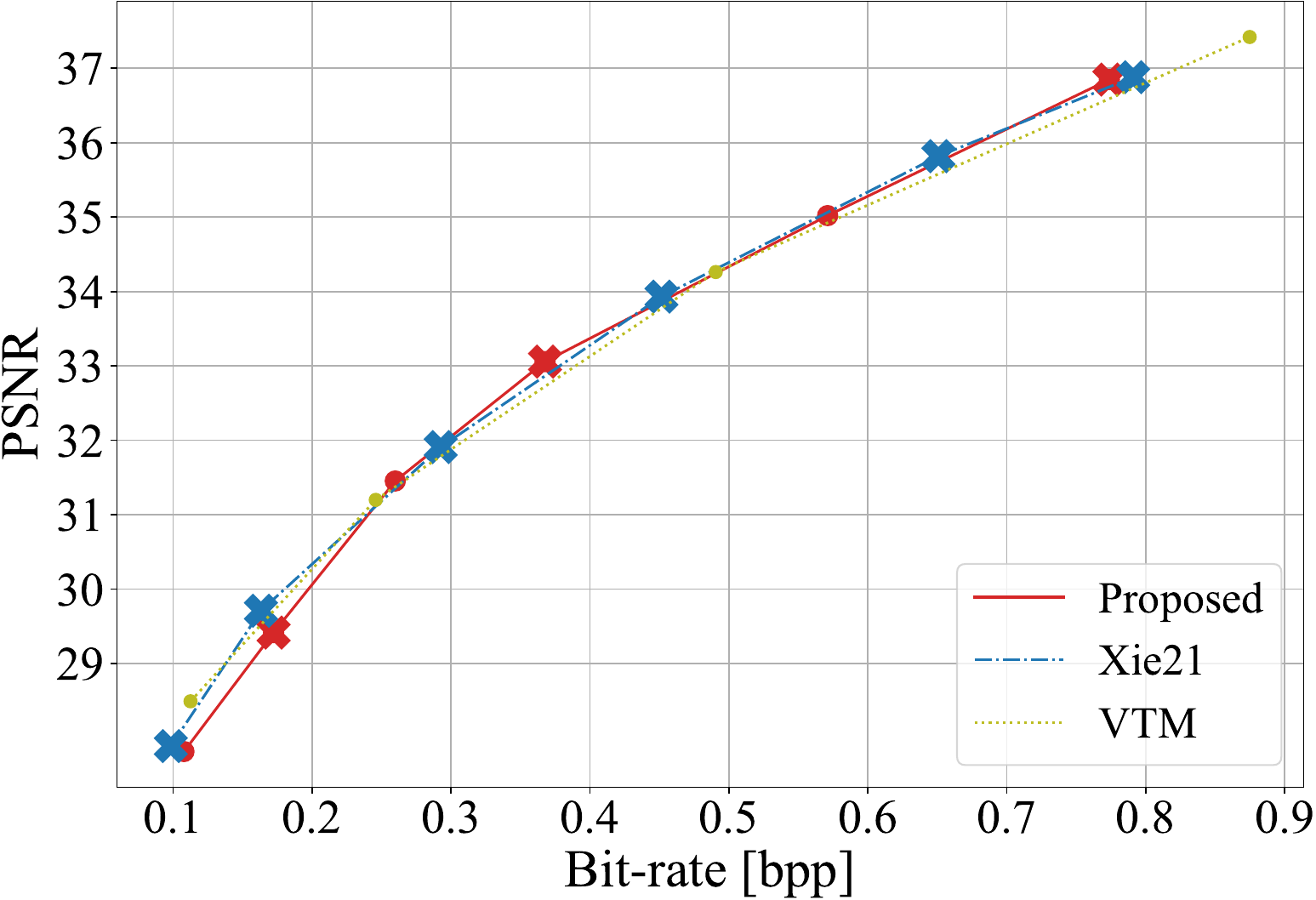}
        \caption{\emph{Xie21}}
        \label{xie2021psnrkodak}
  \end{subfigure}
  \hfill 
    \begin{subfigure}{0.27\textwidth}
        \centering
        \includegraphics[width=\textwidth]{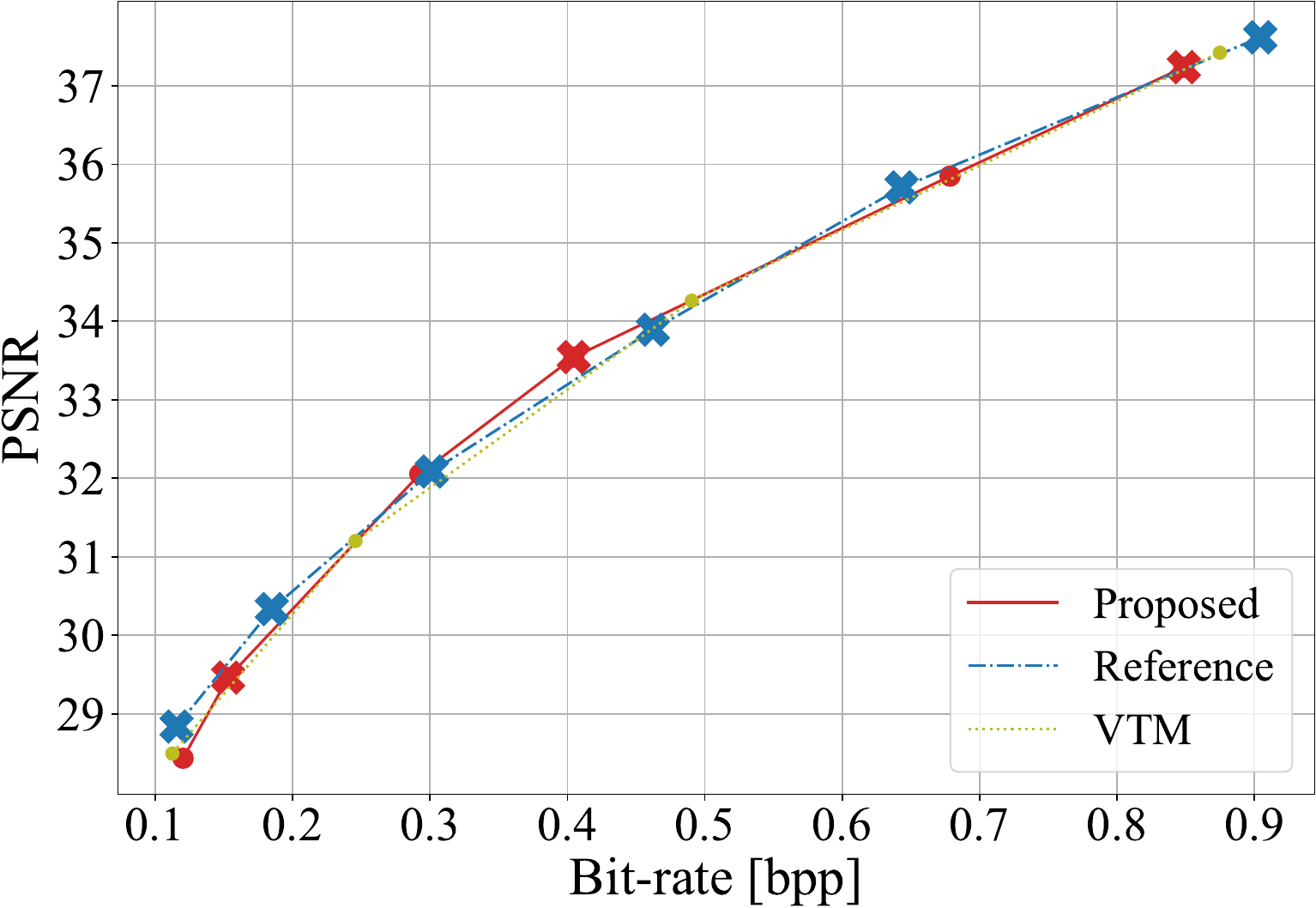}
        \caption{\emph{Zou22}}
        \label{zou2022psnrkodak}
  \end{subfigure}

  \vspace{\baselineskip}

      \begin{subfigure}{0.27\textwidth}
        \centering
        \includegraphics[width=\textwidth]{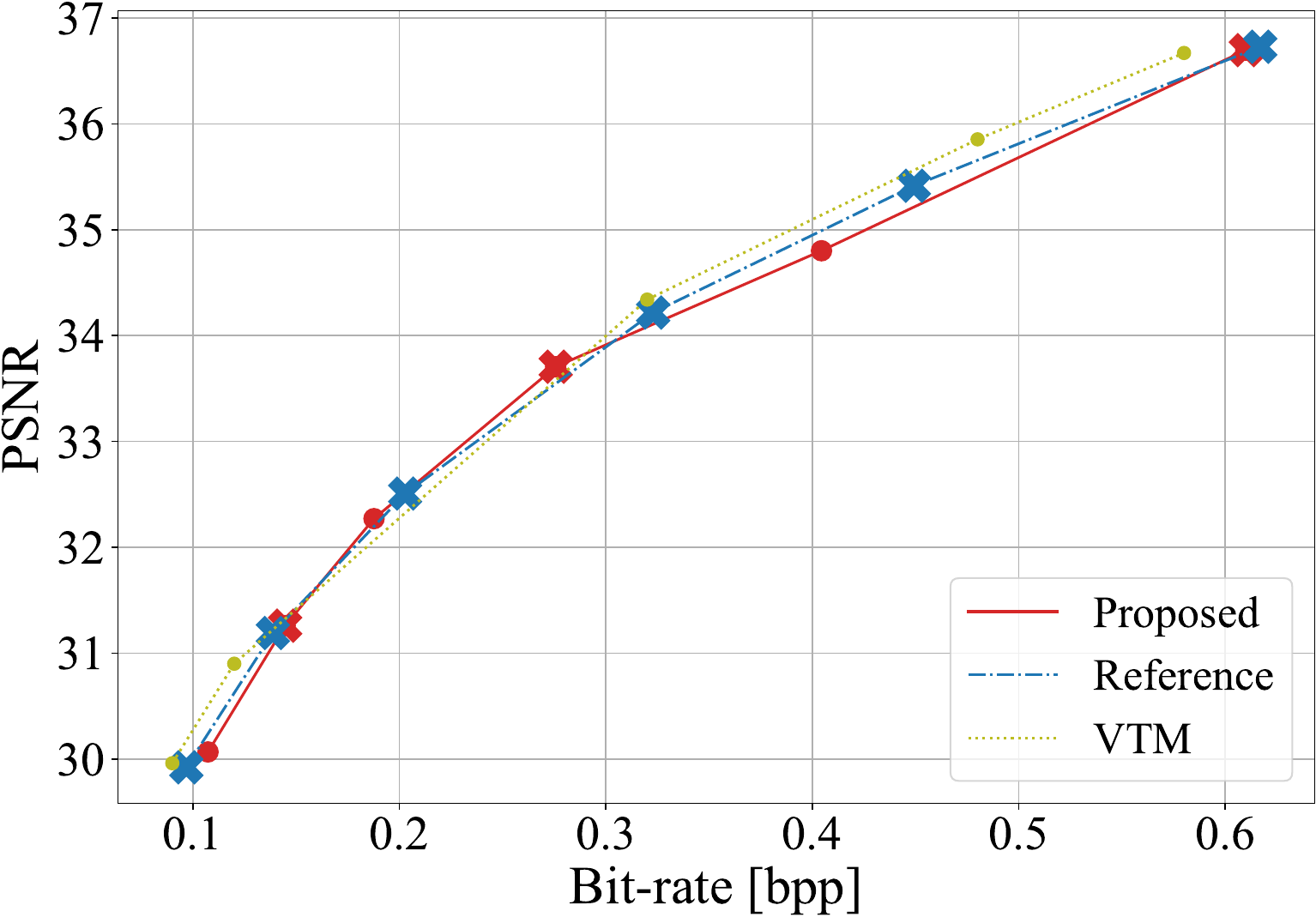}
        \caption{\emph{Cheng20}}
        \label{cheng2020psnrclic}
  \end{subfigure}
  \hfill
    \begin{subfigure}{0.27\textwidth}
        \centering
        \includegraphics[width=\textwidth]{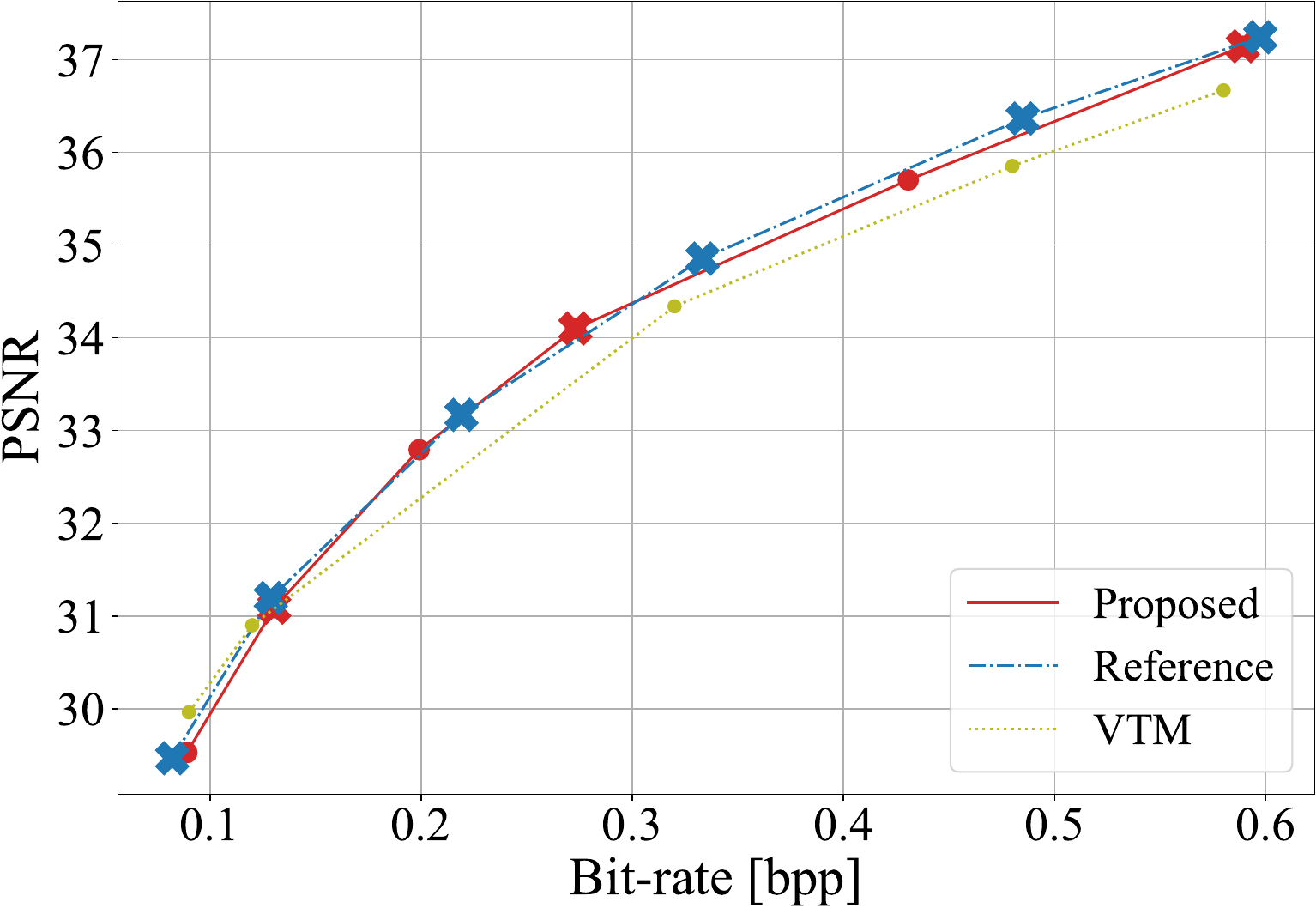}
        \caption{\emph{Xie21}}
        \label{xie2021psnrclic}
  \end{subfigure}
  \hfill 
    \begin{subfigure}{0.27\textwidth}
        \centering
        \includegraphics[width=\textwidth]{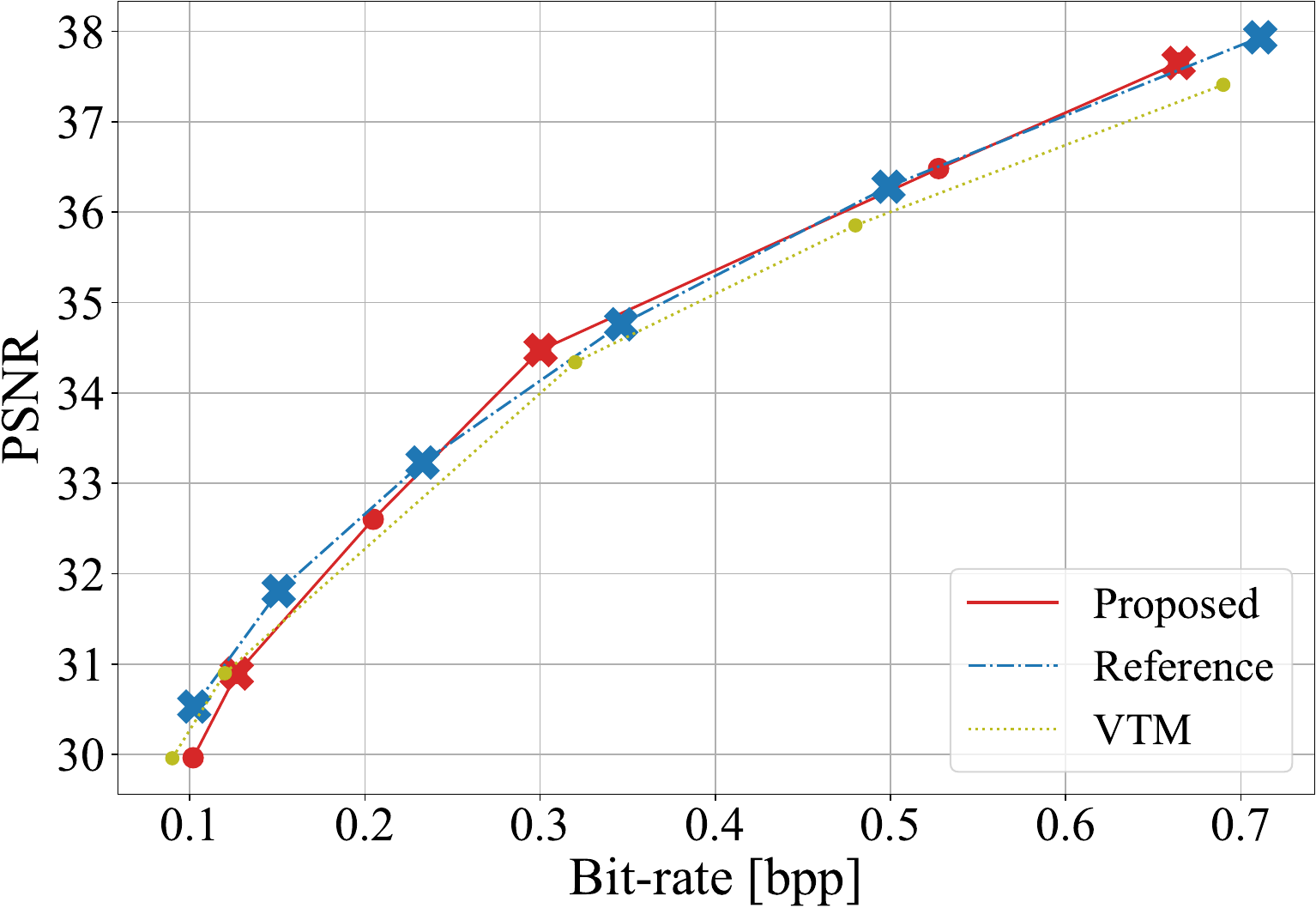}
        \caption{\emph{Zou22}}
        \label{zou2022psnrclic} 
    \end{subfigure}

  \vspace{\baselineskip}
  
  \label{fig:immagini}

      \begin{subfigure}{0.27\textwidth}
        \centering
        \includegraphics[width=\textwidth]{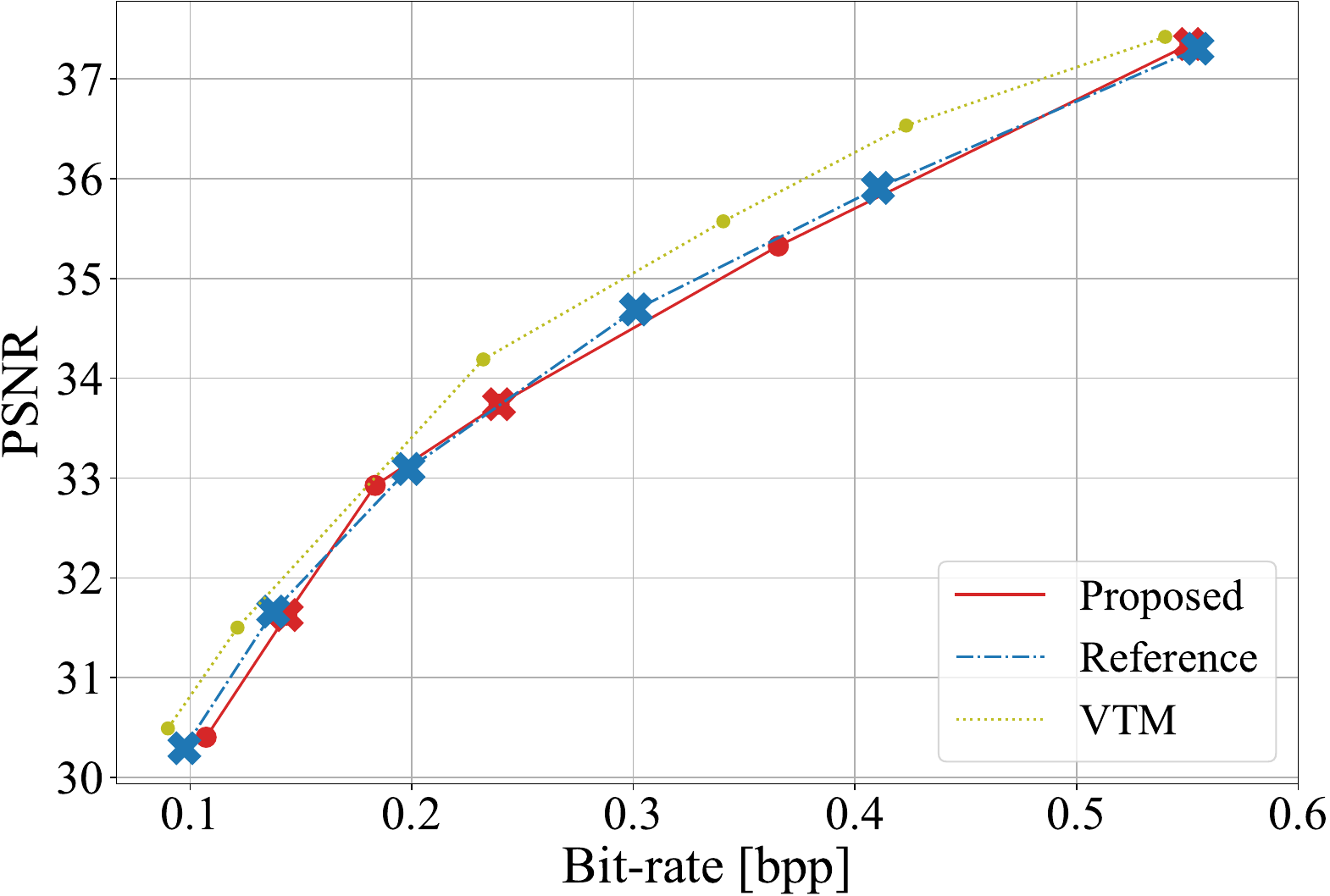}
        \caption{\emph{Cheng20}}
        \label{cheng2020psnrteck}
  \end{subfigure}
  \hfill
    \begin{subfigure}{0.27\textwidth}
        \centering
        \includegraphics[width=\textwidth]{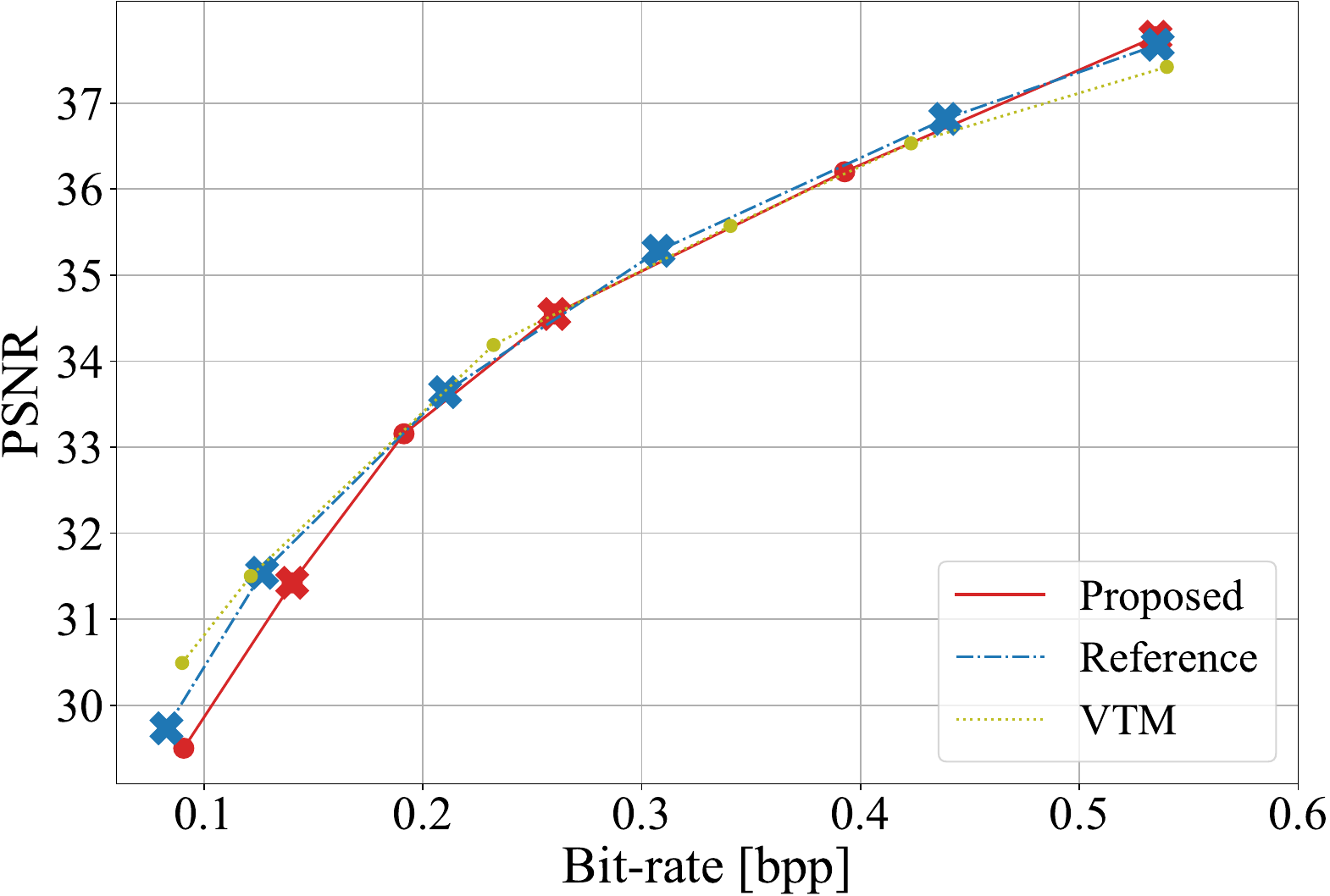}
        \caption{\emph{Xie21}}
        \label{xie2021psnrteck}
  \end{subfigure}
  \hfill 
    \begin{subfigure}{0.27\textwidth}
        \centering
        \includegraphics[width=\textwidth]{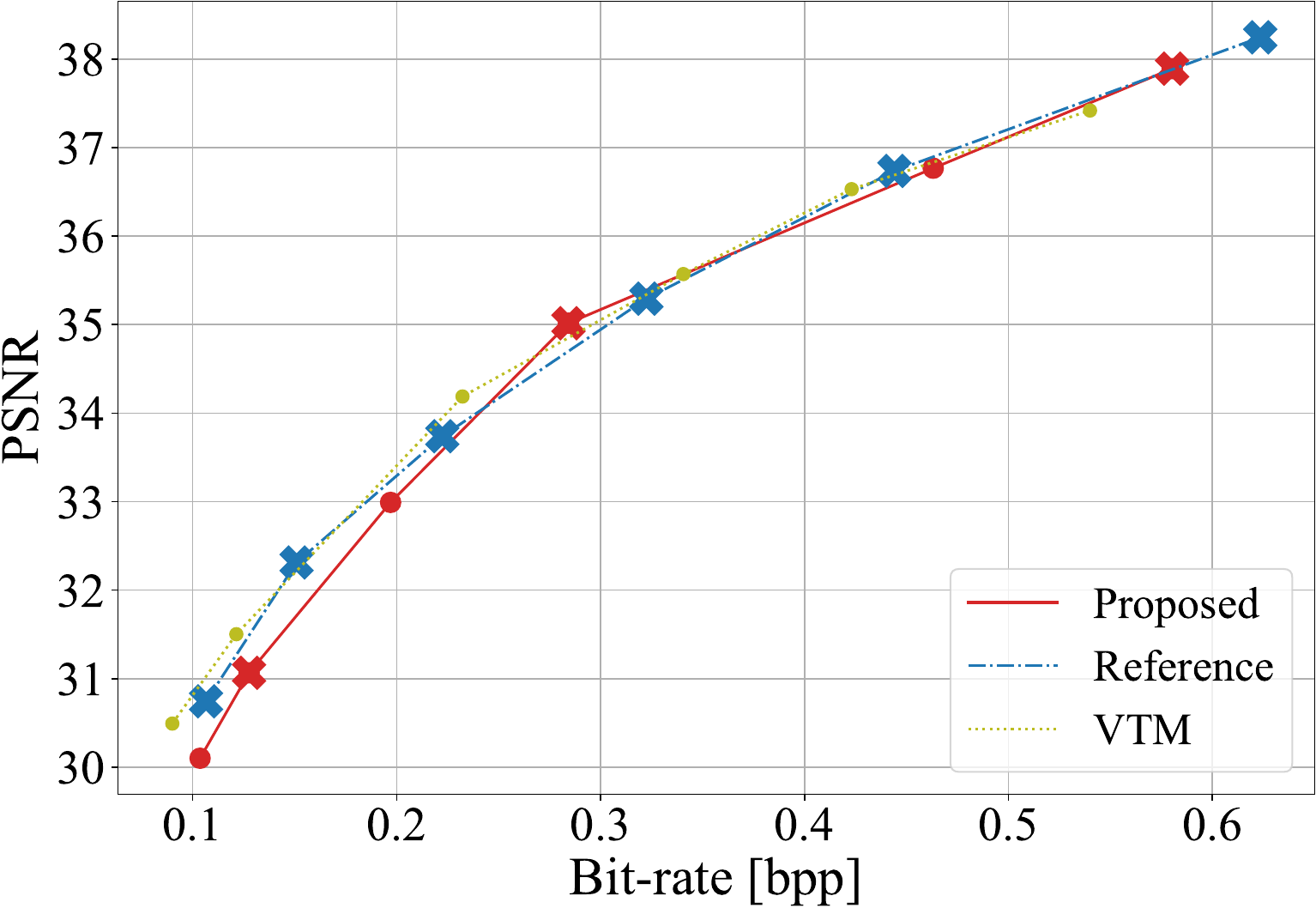}
        \caption{\emph{Zou22}}
        \label{zou2022psnrteck} 
    \end{subfigure}
  \caption{Rate-PSNR plots for the proposed \stanh-based method and relative reference for Kodak (top row), Clic (central row), and Tecnik (bottom row) datasets and for 3 anchors and 3 derivations.} 
  \label{fig:4}
\end{figure*}

\begin{figure}
    \centering
    \includegraphics[width=0.60\columnwidth]{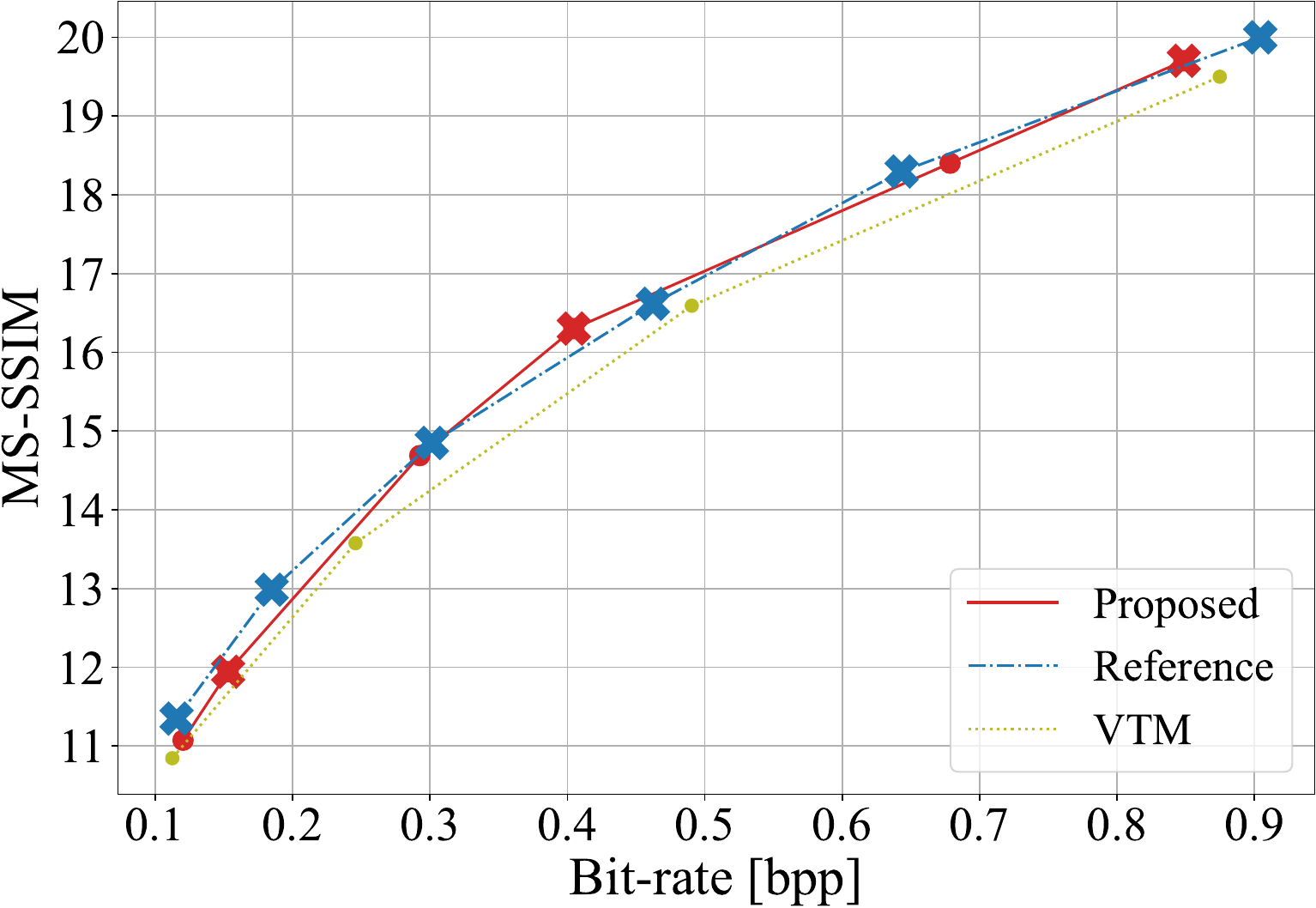}
    \caption{MS-SSIM  for the proposed \stanh-based method on \emph{Zou22}, for 3 anchors and 3 derivations.}
    \label{fig:5}
\end{figure}

\vspace{-0.5cm}

\subsection{Rate-distorsion Performance} \label{performance}

We now extend our experiments to the 
\textit{Xie21} and \textit{Cheng20} while keeping the number of anchors equal to 3: Fig.~\ref{fig:4} shows the three datasets.
As an additional reference, we add a curve for the H.266/VVC reference encoder VTM-20.2~\cite{vvc}.
Our proposed method exhibits curves that overlap almost entirely with the reference curves, i.e. it does not affect the compression efficiency despite a 50\% reduction in complexity.
Similar results can be observed  also in Fig.~\ref{fig:5} where image quality is measured in MS-SSIM terms (we convert \emph{MS-SSIM} to $-10 \log_{10}(1 - \textit{MS-SSIM})$): in reason of space, we report only results related to Kodak dataset on \emph{Zou22}.
Tab.~\ref{tab_bdrate} compares our 3-anchors proposed models with references in terms of BD-Rate and BD-PSNR on the Kodak dataset 
For both \emph{Zou22} and \emph{Cheng20}, the BD-Rate loss is below $1\%$, while for \emph{Xie21} it remains well below $2\%$ despite a reduction in complexity below 50\%, as we discuss in detail later on.

\begin{table}
\centering
\small
 \caption{BD-Rate and BD-PSNR  on Kodak  for 3 anchors,  savings are reported in terms of trainable parameters (TP).   }
\resizebox{\columnwidth}{!}{\begin{tabular}{p{0.21\linewidth}ccccc} 
  \toprule
  \textbf{\emph{Model}} & \textbf{\emph{BD-Rate}}& \textbf{\emph{BD-PSNR}}& \textbf{\emph{TP-proposed}} & \textbf{\emph{TP-reference}} &  \textbf{\emph{Savings} (\%)} \\
  \midrule
  \emph{Cheng20}& 0.97 & -0.02 & 72.8 M & 132.4 M & 45\\
  \emph{Xie21}& 1.72 &  -0.09 & 109.2 M & 244.2 M & 55\\
  \emph{Zou22}& 0.99 & -0.07 & 225.7 M & 451.4 M & 50 \\

  \bottomrule
\end{tabular}}\label{tab_bdrate}
\end{table}

%% file: chapters/4_experiments/3_training_cost.tex
\begin{table*}[!hb]
\caption{Approximate costs in terms of Trainable parameters (TP), Training Time (TT), Training Cost (TC), and Storage (SG). \emph{Reference-total} and \emph{Proposed-total} refers to the cost for training the 6 models covering the entire BD range considered. In parentheses is reported the percentage gain of that particular field compared to the reference.  } 

\centering
\resizebox{\textwidth}{!}{
\begin{tabular}{l   l l l l | l l l l | l l l l  }
\toprule
 \multicolumn{1}{c}{}  & \multicolumn{4}{c|}{\textbf{\emph{Cheng20}}}  & \multicolumn{4}{c|}{\textbf{\emph{Xie21}}} & \multicolumn{4}{c}{\textbf{\emph{Zou22}}} \\
\cmidrule(l){2-5} \cmidrule(l){6-9} \cmidrule(l){10-13}
 & $\sim$TP   &   $\sim$TT (h.)   & 
$\sim$TC (kWh)   & $\sim$ SC  & $\sim$TP      &  $\sim$TT (h.)   & 
$\sim$TC (kWh) & $\sim$SC  & $\sim$TP   &   $\sim$TT (h.)   &
$\sim$TC (kWh) & $\sim$SC  \\
\midrule
\emph{Reference-single model} & 29.6 M   & 96  & 27.7  & 0.12 GB & 50 M    & 188   & 52.5 & 0.2 GB & 75.2 M   & 163   & 42.4 & 0.29 GB \\
\emph{Proposed-anchor}  & 29.6 M  & 112  & 30.7 & 0.12 GB & 50 M  & 198  & 54.6  & 0.2 GB & 75.2 M   & 171 & 44.1 & 0.29 GB \\
\emph{Proposed-derivation} &  240   & 12    &3.4 & 2.5 kB & 240   &  19    &5.3 & 2.5 kB & 320   & 11 & 2.7 & 3 kB \\
\emph{Reference-total} & 132.4 M  & 504    & 139.6 & 0.53 GB & 244.2 M  & 1122   & 321.6 & 1.3 GB & 451.4 M & 978 & 254.3 & 1.76 GB  \\
\emph{Proposed-total} &  72.8 M & 371  & 91.1 & 0.27 GB (49 \%) & 109.2 M  & 584  & 176.9 & 0.56 GB (57 \%) & 225.7 M & 546 & 140.1 & 0.86 GB (53 \%) \\
\bottomrule
\end{tabular} \label{tab_training_cost}
}
\end{table*}
\vspace{-0.25cm}
\subsection{Training cost} \label{complexity}

A largely overlooked aspect of LIC is the cost of training the models from scratch, where a single training can take up to 10 days~\cite{compressai} for a stable solution.
Conversely with \stanh, when refining an anchor into a derivation as detailed in Sec.~\ref{training_procedure}, only the parameters in the quantization layer(s) need to be updated, for less than  $0.001\%$ of the total model complexity in most cases.
We further quantify the training costs in terms of parameters to train or refine (TP), training time (TT), and energy consumption (TC).
For the reference models, we report the numbers in the original papers; otherwise, we train the models to produce the required numbers. 
For \stanh, we consider the usual scheme where we train three anchors from scratch and refine three derivations. Tab. \ref{tab_training_cost} shows that the cost of refining a derivation is just one-tenth of training the reference model.
In fact, refining anchor amounts to training 240-360 M (depending on the model) parameters for each \stanh layer only, rather than a deep convolutional model.
As a result, \stanh saves from 33\% (\textit{Cheng20}) to 45\% (\textit{Zou22}) of the energy required and 48\% of the training time for \textit{Xie21} for training 3 anchors and refining 3 derivations.
Further savings can be of course achieved by replacing further anchors with derivations at the price of somewhat lower RD performance (see Fig.~\ref{fig:3}).
\\
We observed that annealing $\beta$ to a stable configuration increases the training time.  Additionally, the bounds of the integral \eqref{gaussian_rate} vary at each iteration depending on the parameters of \stanh, and calculating these parameters can lengthen the training.
Optimizing this implementation aspect could yield further complexity savings that we leave for our future research, as our method already enables consistent gains considering the total energy consumption and training time.
We specify that these calculations are based on the training information given by the original papers, using our computational resources to calculate the average power consumption and multiplying it by the hours required for the network training: the reported energy value is therefore an estimation of the real value; however, it shall allow appreciating a drastic improvement when refining the \stanh layer only.

%% file: chapters/4_experiments/4_memory_usage.tex
\vspace{-0.25cm}
\subsection{Storage cost} \label{memory}

Another overlooked aspect of LIC is the requirements for storing the trained models on user devices,
especially resource-constrained devices such as mobiles, settop-boxes, SoCs, etc., where storage is limited by design.
Tab. \ref{tab_training_cost} presents (fourth column \textit{SC}) the storage requirements for the \textit{pickle} format, and we recall the footprint of the models varies depending on the model and the size of the latent spaces.
The reference schemes need to store 6 models, whereas our proposed method needs to store 3 (or fewer) anchors.
We recall that storing the derivations amounts to storing only the few hundred parameters of the refined \stanh layers.
The storage cost reduction is about 50\%: for \emph{Cheng20} and \emph{Zou22} we report 49\% and 53\% savings, respectively.
The most significant improvement is observed with \emph{Xie21}, where the footprint is reduced by $\sim 57\%$. This is because the 2 lowest bitrate anchors are smaller in this case (with $N = 128$).
We hypothesize that such numbers could be further reduced if the models were saved in some compressed format, albeit this goes out of the scope of this work.

%% file: chapters/4_experiments/5_comprehensive_result_spiders.tex
\vspace{-0.25cm}
\subsection{Comparison with variable rate models} \label{sec:comp_vr_models}
\begin{figure}[t!]
  \centering
 \includegraphics[width=1.0\columnwidth]{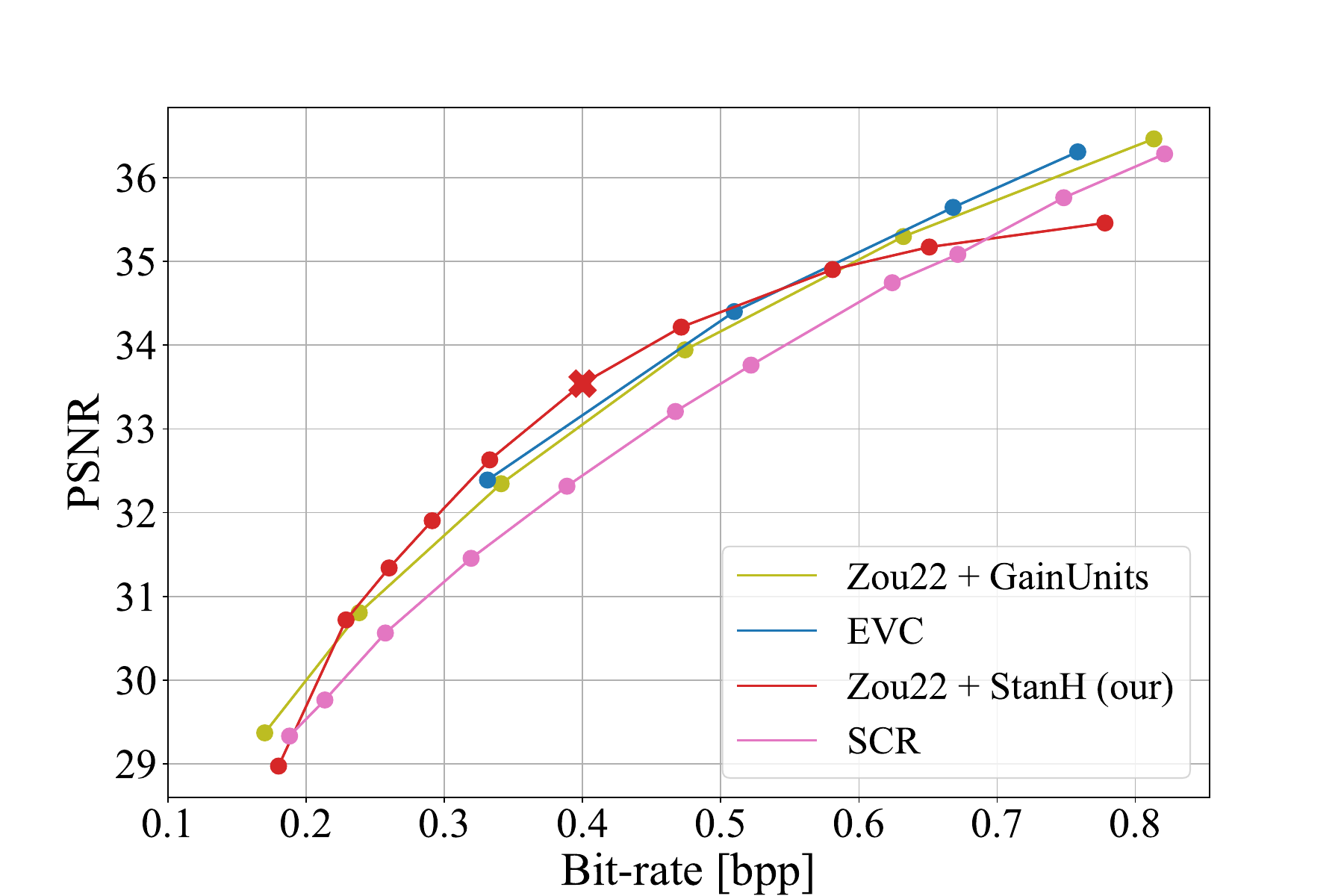}
  \caption{RD-plot on  Kodak  for the proposed \stanh-based method vs. \emph{Gain} \cite{gain21} over \emph{Zou22}, \emph{EVC} \cite{evc2023} and \emph{SCR} \cite{Jooyoung2022}.}
  \label{fig:variable}
\end{figure}

In this section, we compare \stanh with \emph{Gain} \cite{gain21} ,\emph{EVC} \cite{evc2023}, and \emph{SCR} \cite{Jooyoung2022} three state-of-the-art VBR codecs we introduced in Sec.~\ref{related}.
About \emph{EVC} and \emph{SCR}, we  take as reference the numbers from the original papers, as both rely on ad-hoc architectures and training procedures.
Regarding \emph{Gain}, it achieves adaptive quantization plugging into a model, like \stanh, yet it relays on an ad-hoc entropy model distribution of the latent representation.
For a fair comparison, we implemented gains units ourselves over the same \emph{Zou22} model we took as a reference for \stanh.
Training is performed according to the process described in the original \emph{Gain} paper, using six different qualities.
Furthermore, we used only one anchor (A2 in Sec.\ref{performance}) to cover the entire range, since \cite{gain21} is composed by only one model. 
\\
Fig.~\ref{fig:variable} shows that \stanh outperforms the three references in proximity of the A2 (0.25-0.50 bpp range).
When moving away from the anchor, the RD efficiency of \stanh  degrades gracefully, 
and Tab. \ref{tab_var} shows that \stanh is still the best performer (minimal RD efficiency loss) over a  H.266/VVC \cite{vvc} reference software VTM . 
In terms of complexity, \stanh and \emph{Gain} have similar training costs, as only one anchor needs to be trained from scratch.
Also the storage costs is comparable since both methods need to store only one anchor plus the few thousands extra parameters that enable VBR coding.
Regarding \emph{EVC}, it introduced a dual prior encoder that extrapolate a point-wise gain unit in order to obtain bitstreams at different quality. Because of this, it necessitates more parameters to obtain VBR (\ref{tab_var}), moreover this method is not agnostic with respect to model architecture (e.g., it is not possibile to use as it is on channel-wise model like \emph{Zou22}).

\begin{table}
\centering
\small
 \caption{BD-Rate, BD-PSNR, Number of parameters for variable bitrate (\#Pars for VBR), and storage cost (SC) in GB of \stanh with one anchor vs. \emph{Gain} vs. \emph{EVC} over a H.266/VVC reference.}
\resizebox{1\columnwidth}{!}{\begin{tabular}{p{0.3\linewidth}cccc} 
  \toprule
  \textbf{\emph{}} & \textbf{\emph{BD-Rate}}& \textbf{\emph{BD-PSNR}}& \textbf{\emph{\#Pars for VBR}} & \emph{SC}  \\
  \midrule
  \emph{Zou22 + }\stanh & \textbf{0.82} & \textbf{-0.15} &  \textbf{3200} & 0.29 \\
  \emph{Zou22 + Gain} & 5.6 & -0.29 &  6144 & 0.29\\
    \emph{EVC} & 0.96 & -0.16 &  1663108 & 0.116 \\
    \emph{SCR} & 18.72 & -0.85 &  69440 & \textbf{0.05} \\

  \bottomrule
\end{tabular}}\label{tab_var}
\end{table}

\subsection{Continuous rate adaptation}\label{overview}

\begin{figure}[h!]
  \centering
 \includegraphics[width=1.0\columnwidth]{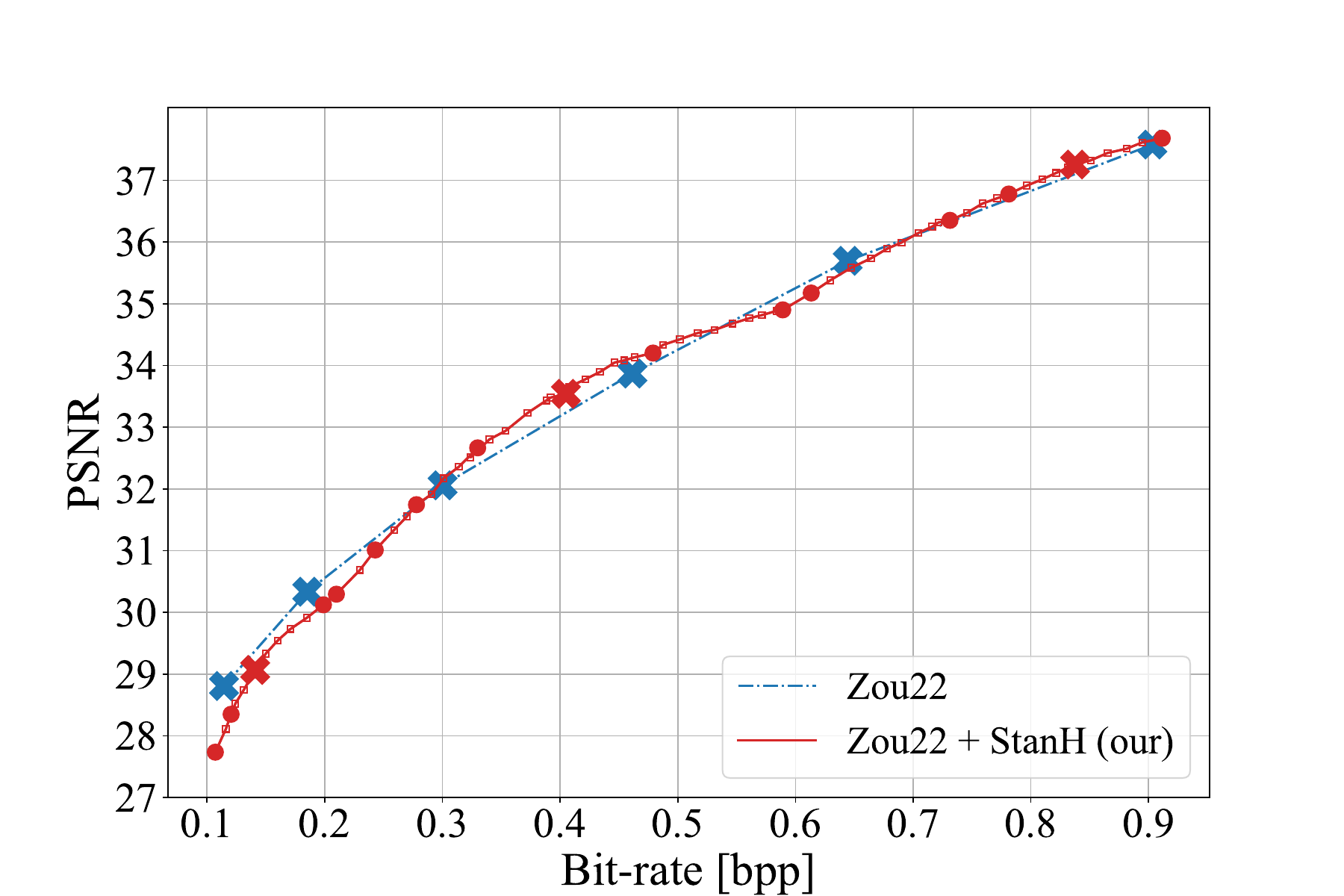}
  \caption{ RD-plot for \stanh-based model and the corresponding reference for Kodak, considering \emph{Zou22}. Stars represent anchor models, circles tuned \stanh's , and small empty squares the interpolations.}
  \label{fig:6}
\end{figure}

\begin{figure}[h!]
  \centering
 \includegraphics[width=0.55 \columnwidth]{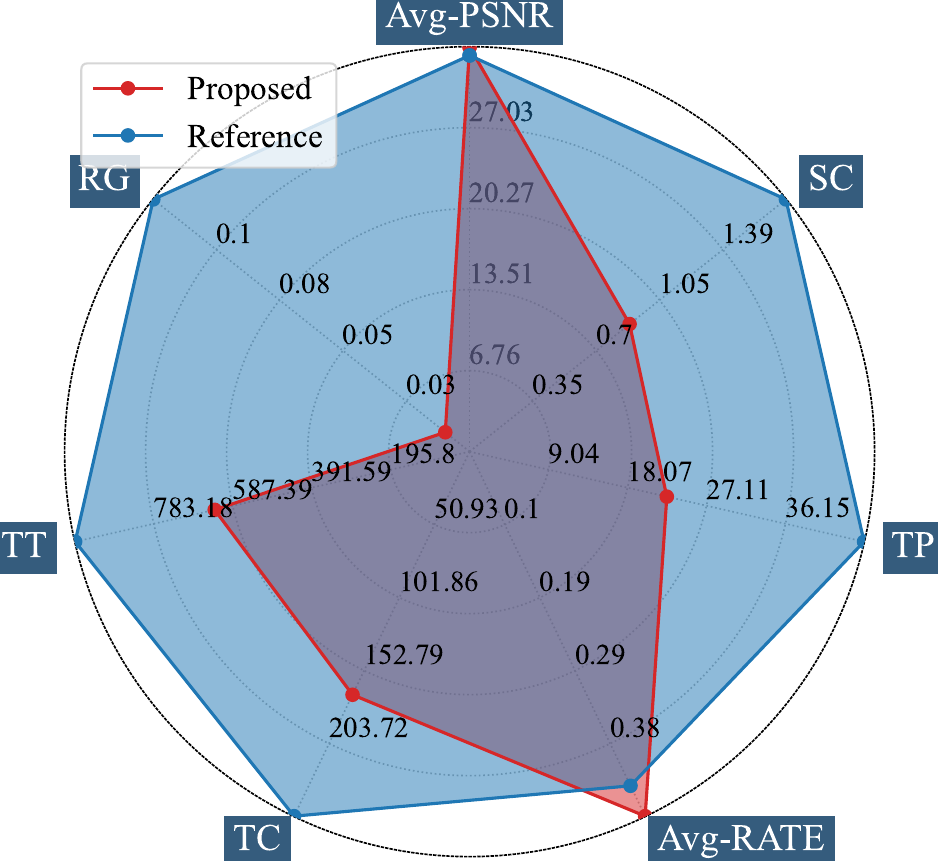}
  \caption{Radar plot comparing the reference and proposed schemes in terms of average PSNR (\emph{Avg-PSNR}) and average bitrate (\emph{Rate}), \emph{storage cost} in GB (SC),  number of \emph{trainable parameters} in millions ($TP$), \emph{training cost} in kWh ($TC$), \emph{training time} in hours (TT), and rate granularity in bpp (\emph{RG}).}
  \label{fig:7}
\end{figure}

In this section we evaluate the ability of our method to achieve both  fine-grained and continuous rate control using the interpolation strategy described in Sec. \ref{fctf}.
From the 3 initial anchors, we tuned a total of 13 derivations (4160 additional parameters), which have been exploited to interpolate about 50 extra RD points, i.e. interpolations, to achieve continuous rate adaptation.
Fig.~\ref{fig:6} shows the resulting RD curves for the \emph{Zou22} architecture on the Kodak dataset, where stars represents the anchors, the circles the derivations and the empty boxes are the interpolations. 
The suggests that interpolating \stanh layers by sweeping $\rho$ in eq. \ref{fctf} affect minimally the RD performance, allowing to achieve continuous variable rate. 
\\
Fig.~\ref{fig:7} condenses in a single radar plot for \stanh and the reference schemes 7 different metrics, namely average PSNR (\emph{Avg-PSNR}) and average bitrate (\emph{Avg-Rate}), \emph{storage cost} in GB (SC),  number of \emph{trainable parameters} ($TP$), \emph{training cost} in kWh ($TC$), \emph{training time} in hours (TT), and rate granularity (\emph{RG}), defined as the average rate distance between adjacent RD points.Apart from \emph{Avg-PSNR}, for all these metrics \emph{the lower, the better}, i.e. a narrower radar profile corresponds to better performance. 
The plot confirms that \stanh enables almost identical RD performance yet for lower training and storage costs, improving all the considered aspects with respect to \emph{Zou22}, including the benefit of continuous rate granularity.

%% file: chapters/5_conclusion/5_conclusion.tex
\section{Conclusions and future works} \label{conclusion}

We proposed a novel method to convert fix-rate LIC models to variable rates by exploiting \stanh, a parametric module that approximates quantization. 
By definition, \stanh converges to the stepwise quantizer if its inverse temperature $\beta$ is properly annealed at training time.
We achieve variable rate by training only a few anchor models end-to-end, and then refining the \stanh layers only for other RD tradeoffs into different derivation models.
Once the anchors have been trained, refining additional derivations has negligible training and storage costs, practically enabling both fine-grained and continuous rate control,  by computing weighted average of already existing \stanh's.
In summary, we show that our method achieves comparable results with respect to both fix-rate and variable-rate LIC models; moreover, thanks to its simplicity it is totally agnostic to the reference architecture. 
\\
In perspective, the goal is to extend our method to learnable video compression: the task is not straightforward since the latent space distributions may differ as the latent spaces often represent residuals, motivating a separate essay on this topic.